\documentclass{article} 
\usepackage{iclr2025_conference,times}

\usepackage{wrapfig}
\usepackage{lipsum} 
\usepackage{wrapfig}

\usepackage{amsmath,amsfonts,bm}
\usepackage{multicol}

\def\eqref#1{equation~\ref{#1}}

\def\1{\bm{1}}

\DeclareMathAlphabet{\mathsfit}{\encodingdefault}{\sfdefault}{m}{sl}
\SetMathAlphabet{\mathsfit}{bold}{\encodingdefault}{\sfdefault}{bx}{n}

\usepackage{hyperref}
\usepackage{url}
\usepackage{xcolor}
\usepackage{microtype}
\usepackage{graphicx}

\usepackage{subfloat}
\usepackage{listings}
\usepackage{xcolor}
\usepackage{booktabs} 
\usepackage{float}
\usepackage{amssymb}
\usepackage{color}
\usepackage{mathtools}
\usepackage{amsthm}
\usepackage{multirow}
\usepackage{cleveref}
\usepackage{caption}
\usepackage{mathtools}
\usepackage{amsthm}
\usepackage{amsmath}
\usepackage{algorithm} 
\usepackage{algorithmicx}
\usepackage{algpseudocode}
\usepackage{setspace} 
\usepackage{wrapfig}
\usepackage{cleveref}
\usepackage{diagbox}
\usepackage{xcolor}
\usepackage{svg}
\usepackage{adjustbox} 
\usepackage{minted}
\usepackage{fancyvrb}
\usepackage[T1]{fontenc}
\usepackage{graphicx}    
\usepackage{subcaption}  
\usepackage{booktabs}    
\usepackage{array}       
\usepackage[table]{xcolor}

\usepackage[most]{tcolorbox} 
\usepackage{xcolor}          

\definecolor{uclablue}{rgb}{0.15, 0.45, 0.68}
\hypersetup{
    breaklinks,
    citecolor=uclablue,
    colorlinks=true,
}

\NewDocumentCommand{\xx}
{ mO{} }{\textcolor{blue}{\textsuperscript{\textit{todo}}\textsf{\textbf{\small[#1]}}}}
\usepackage{xspace}

\definecolor{greenbg}{RGB}{230, 255, 230}

\title{ProAct: Agentic Lookahead in Interactive Environments}

\author{Yangbin Yu*, Mingyu Yang*, Junyou Li, Yiming Gao, Feiyu Liu, Yijun Yang, Zichuan Lin\\ Jiafei Lyu, Yicheng Liu, Zhicong Lu, Deheng Ye, Jie Jiang$\dagger$\\
Tencent Hunyuan\\
}

\begin{document}
\maketitle
\renewcommand*{\thefootnote}{\fnsymbol{footnote}}
\footnotetext{* Equal contribution.}
\footnotetext{$\dagger$ Corresponding Author. (Email:zeus@tencent.com)}

\begin{figure}[h]
\centering
\includegraphics[width=1\textwidth]{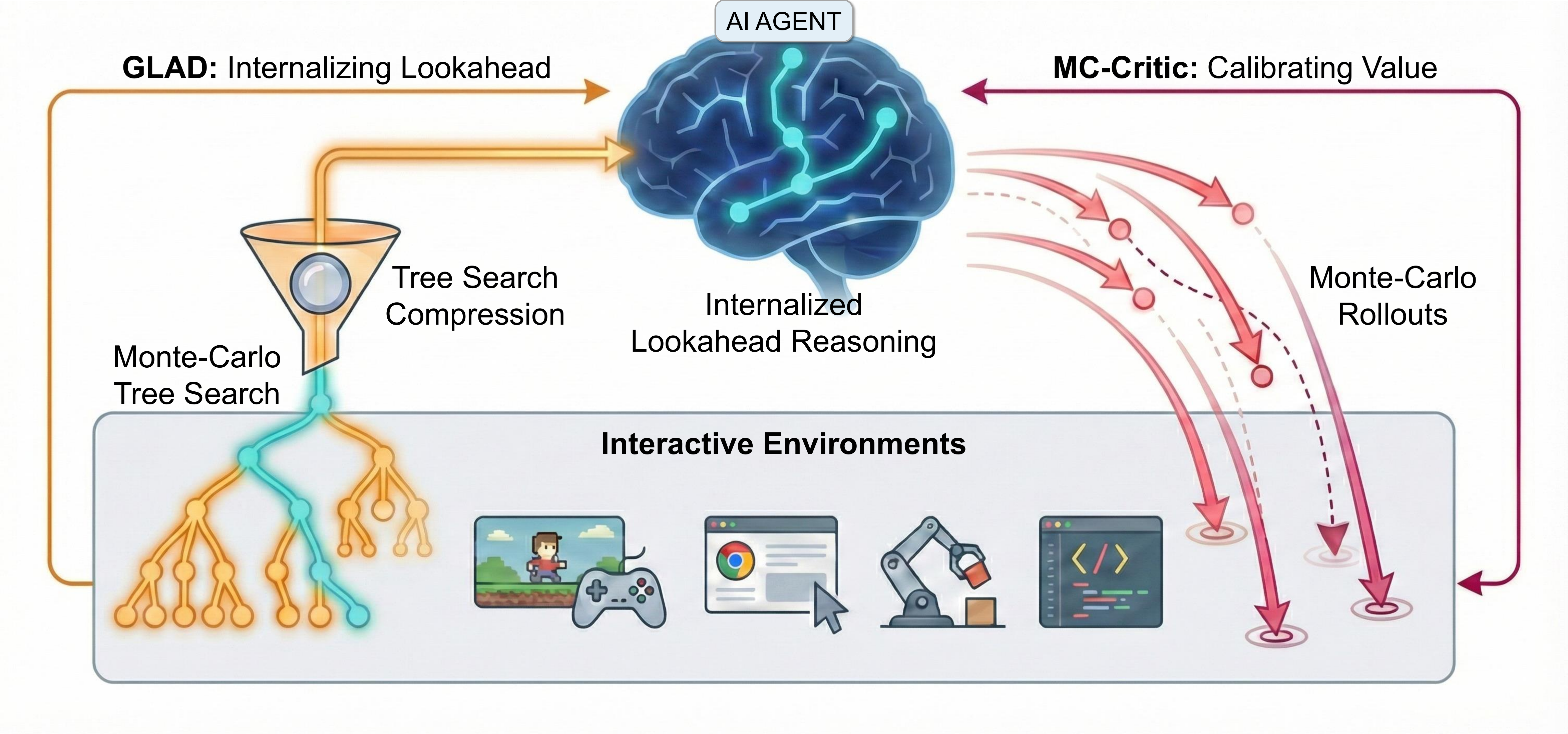}

\caption{
Overview of \textbf{ProAct}. A two-stage paradigm to internalize accurate lookahead reasoning for AI agents. \textbf{GLAD} distills complex MCTS search trees into concise, causal reasoning chains via SFT. \textbf{MC-Critic} leverages lightweight environment rollouts to provide low-variance value estimates, stabilizing online RL training.
}

\vspace{2em}
\label{fig:proact_framework}
\end{figure}

\begin{abstract}
Existing Large Language Model (LLM) agents struggle in interactive environments requiring long-horizon planning, primarily due to compounding errors when simulating future states. To address this, we propose \textbf{ProAct}, a framework that enables agents to internalize accurate lookahead reasoning through a two-stage training paradigm. First, we introduce Grounded LookAhead Distillation (\textbf{GLAD}), where the agent undergoes supervised fine-tuning on trajectories derived from environment-based search. By compressing complex search trees into concise, causal reasoning chains, the agent learns the logic of foresight without the computational overhead of inference-time search. Second, to further refine decision accuracy, we propose the Monte-Carlo Critic (\textbf{MC-Critic}), a plug-and-play auxiliary value estimator designed to enhance policy-gradient algorithms like PPO and GRPO. By leveraging lightweight environment rollouts to calibrate value estimates, MC-Critic provides a low-variance signal that facilitates stable policy optimization without relying on expensive model-based value approximation. Experiments on both stochastic (e.g., 2048) and deterministic (e.g., Sokoban) environments demonstrate that ProAct significantly improves planning accuracy. Notably, a 4B parameter model trained with ProAct outperforms all open-source baselines and rivals state-of-the-art closed-source models, while demonstrating robust generalization to unseen environments. The codes and models are available at \url{https://github.com/GreatX3/ProAct}.

\end{abstract}



\section{Introduction}
Recent advancements in Large Language Models (LLMs) have enabled the development of autonomous agents capable of performing complex tasks ~\citep{agent-survey,agentgym-rl,llm-agent-surve-challenges}. Unlike static question-answering, interactive environments require agents to make sequential decisions where current actions significantly constrain future possibilities. In such settings, human cognition typically relies on System 2 processing ~\citep{openai-o1,system1,deepseek-r1}: rather than acting greedily on the current state, humans perform mental lookahead—simulating potential future trajectories and comparing outcomes before committing to an action.

However, replicating this lookahead capability in LLM agents remains a fundamental challenge. While techniques like Chain-of-Thought ~\citep{cot,tot} and ReAct ~\citep{react} have improved reasoning, they suffer from a critical limitation in long-horizon tasks: compounding simulation errors ~\citep{agent-hallucinations}. When an LLM  agent attempts to simulate future states, minor inaccuracies in predicting the dynamics of the environment accumulate rapidly ~\citep{agent-hallucinations,llm-agent-surve-challenges}. Consequently, the reasoning process, though logically sound in structure, diverges from reality, leading to delusional plans and suboptimal decision-making. Simply increasing the depth of ungrounded reasoning often further degrades performance due to context drift and hallucination, while single-step reasoning fails to capture long-term consequences.

To bridge this gap, we propose ProAct, a framework designed to internalize accurate, multi-turn lookahead capabilities into LLM agents. Our core insight is that while LLMs struggle to generate accurate simulations zero-shot, they can effectively learn the pattern of correct simulation if provided with grounded supervision. ProAct operates in two stages as shown in Figure~\ref{fig:proact_framework}. First, during Supervised Fine-Tuning (SFT), we employ Grounded LookAhead Distillation (GLAD). We utilize the environment to perform Monte-Carlo Tree Search (MCTS), exploring multiple future trajectories including both optimal paths and potential dead-ends. Crucially, instead of training on the raw, expansive search trees, we compress these trajectories into concise, high-quality reasoning chains. By explicitly training on these compressed reasoning chains, the model learns to predict the outcomes of candidate actions and identify risks based on ground-truth environmental feedback, rather than relying on hallucinated physics. Second, to further refine the precision of this internalized model, we introduce a Reinforcement Learning (RL) phase augmented by a Monte-Carlo Critic (MC-Critic). Standard multi-turn RL methods ~\citep{ppo,grpo,turn-ppo} for LLM agents often struggle with value estimation in long-horizon tasks. Our MC-Critic leverages the interactive nature of the environment to perform lightweight rollouts, providing a suboptimal low-variance estimate of the value function. This allows the model to align its internalized lookahead with the true environmental dynamics efficiently.

To rigorously evaluate ProAct, we utilize long-horizon games (e.g., 2048, Sokoban) as testbeds. These environments provide objective metrics for reasoning capabilities, featuring both stochasticity and strict deterministic dynamics. Experiments show that ProAct significantly enhances multi-turn decision-making capabilities. Notably, a 4B parameter model trained with our framework outperforms all open-source baselines and achieves performance comparable to state-of-the-art closed-source models, while exhibiting strong generalization to unseen environment configurations.

Our contributions are summarized as follows:
\begin{enumerate}
    \item We propose \textbf{GLAD}, a method to internalize environmental dynamics into LLM agents by compressing search-based trajectories into explicit reasoning chains, mitigating simulation hallucinations.
    \item We introduce \textbf{MC-Critic}, a plug-and-play auxiliary value estimation method that leverages environment interaction to stabilize and accelerate multi-turn agentic RL training.
    \item We demonstrate that a 4B parameter model trained with \textbf{ProAct} outperforms all open-source baselines and achieves performance comparable to state-of-the-art closed-source models. Furthermore, the model demonstrates strong generalization to unseen environments.
\end{enumerate}

\section{Related Work}
Recent research in LLM agents has shifted from static, single-turn problem solving to dynamic, multi-turn interactions in complex environments. This paradigm shift necessitates advancements across three critical dimensions: (1) Multi-Turn Agentic RL, which addresses the stability and exploration challenges inherent in trajectory-level optimization; (2) Reasoning and Search Distillation, where expensive inference-time planning (System 2) is compressed into efficient policy intuition (System 1); and (3) Value Estimation for Agentic RL, which tackles the difficulty of attributing sparse rewards to specific reasoning steps in long-horizon tasks. Our work synthesizes these directions, proposing a unified framework that calibrates agent reasoning via environment-based search and stabilizes policy updates through Monte-Carlo value estimation.

\paragraph{Multi-Turn Agentic Reinforcement Learning}
Training LLMs as agents requires moving beyond single-turn alignment to trajectory-level optimization. Recent frameworks like AgentGym-RL ~\citep{agentgym-rl} and RAGEN ~\citep{ragen} have established the foundations for this transition. AgentGym-RL ~\citep{agentgym-rl} introduces ScalingInter-RL, a curriculum strategy that progressively expands interaction horizons to prevent model collapse during exploration. Similarly, RAGEN ~\citep{ragen} identifies the ``Echo Trap" phenomenon in multi-turn RL where agents overfit to shallow reasoning patterns and proposes StarPO to stabilize trajectory-level updates. Other systems like SkyRL ~\citep{skyrl} and DART ~\citep{dart} focus on the asynchronous execution efficiency of these long-horizon rollouts. While these frameworks ~\citep{uitars-2} provide the infrastructure for agent training, our work focuses on the quality of the internal reasoning process of agent. Unlike methods that rely on implicit rewards or pure trial-and-error, we introduce an environment-calibrated reasoning paradigm that actively mitigates the cumulative errors inherent in long-horizon internal simulations.

\paragraph{Reasoning Distillation from System 2 to System 1}
Enhancing agents with ``System 2" capabilities is a central theme in recent research. Methods like Tree of Thoughts ~\citep{tot} and RAP ~\citep{rap} integrate explicit search algorithms (e.g., BFS, MCTS) during inference to explore reasoning paths. However, the high computational cost of inference-time search limits their scalability ~\citep{ma,aga}. Consequently, research has pivoted towards distillation: transferring the capabilities of expensive planners into efficient ``System 1" policies. Distilling Step-by-Step ~\citep{distill-step-by-step} and STaR ~\citep{star} demonstrate that LLMs can bootstrap their own reasoning capabilities by fine-tuning on self-generated rationales. Furthermore, VAGEN ~\citep{vagen} explicitly enforce the generation of internal ``world models" (e.g., state estimation and dynamics simulation) to ground agent reasoning. Our approach advances this direction by introducing Reasoning Compression. Instead of simply cloning verbose search traces or explicit world model states, we use MCTS to calibrate the agent's thought process against ground-truth environmental dynamics, and then compress this search tree into a concise, natural language estimation of future trends. This effectively distills the foresight of MCTS (System 2) into a token-efficient policy (System 1) that maintains diversity and accuracy without the overhead of explicit search tags.

\paragraph{Value Estimation in Agentic Reinforcement Learning}
Accurate value estimation is critical for guiding exploration in sparse-reward environments, yet it remains a bottleneck for RL agents. Traditional neural critics ~\citep{ppo} often suffer from high bias and slow convergence in high-dimensional language spaces. To address this, ArCHer ~\citep{archer} proposes a hierarchical architecture that estimates value at the utterance level to improve credit assignment over long horizons. SWEET-RL ~\citep{sweet} introduces an asymmetric critic with access to privileged training-time information to provide dense step-level rewards. Turn-Level Reward Design ~\citep{turn-ppo} further explores granular feedback mechanisms to extend algorithms like GRPO ~\citep{grpo} to multi-turn settings. Diverging from these parametric approaches, our method introduces a Monte-Carlo Critic. Instead of training a separate critic network, we leverage low-cost Monte Carlo rollouts to estimate state values dynamically. This provides a low-variance, environment-grounded signal that significantly enhances the stability of policy gradient methods (e.g., PPO ~\citep{ppo}, GRPO ~\citep{grpo}).
\section{The ProAct Framework}
\begin{figure}[h]
\centering
\includegraphics[width=1 \textwidth]{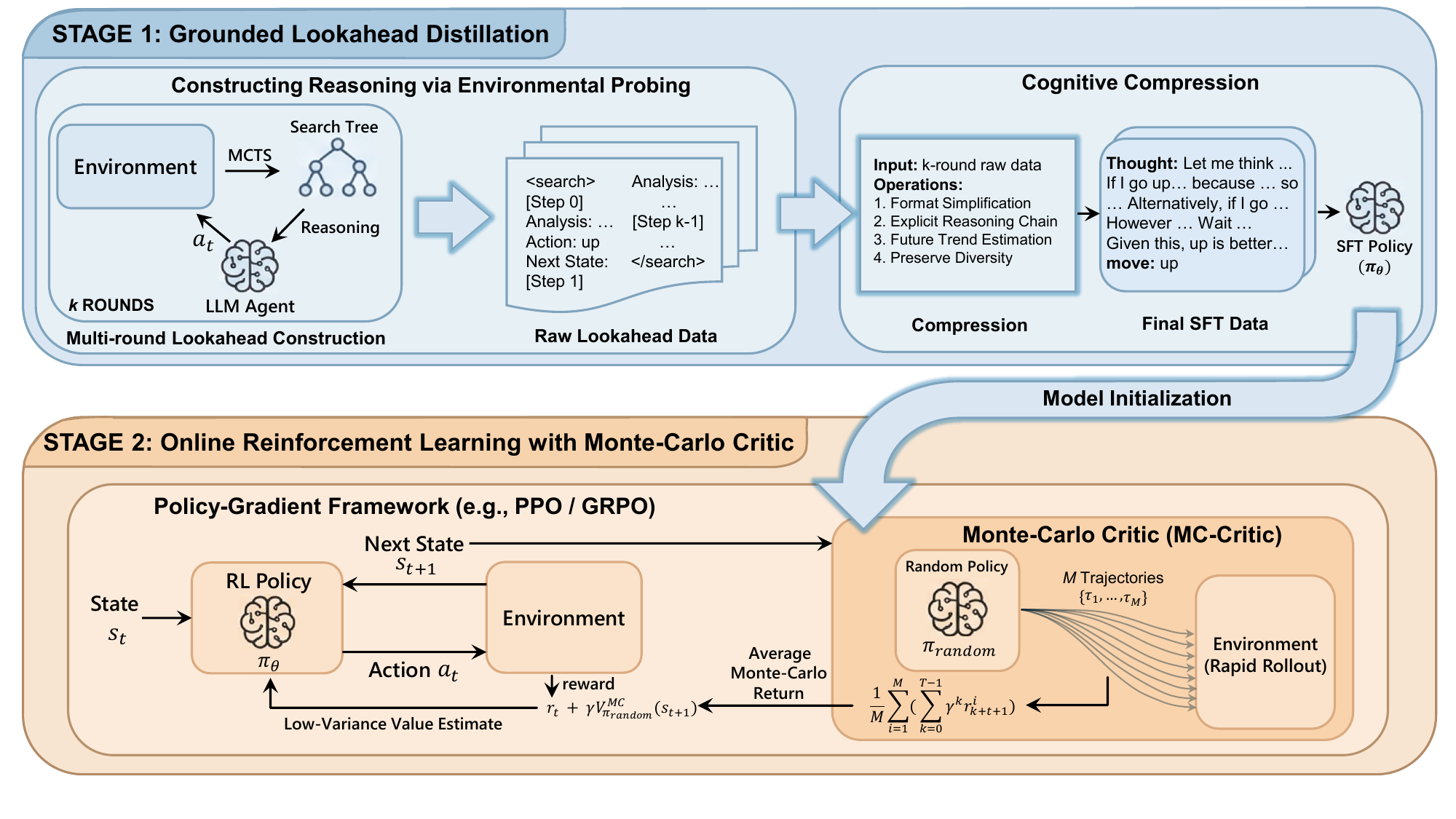}

\caption{
The overall framework of ProAct which operates in two stages to internalize lookahead reasoning capabilities. Stage 1: Grounded Lookahead Distillation establishes the reasoning paradigm. It constructs high-quality lookahead search trees via environmental probing (MCTS) and distills these complex trajectories into compressed, explicit reasoning chains. Stage 2: Online Reinforcement Learning with Monte-Carlo Critic which refines the lookahead reasoning accuracy of agent using a Policy-Gradient framework (e.g., PPO or GRPO). We introduce a plug-and-play MC-Critic that provides low-variance value estimates by aggregating discounted returns from $M$ parallel trajectories generated by a lightweight random policy.
}

\vspace{2em}
\label{fig:proact_overview}
\end{figure}

\subsection{Overview}
\paragraph{Formulation}
We consider the problem of an LLM agent interacting with environments, formalized as a Markov Decision Process (MDP) tuple $\mathcal{M} = \langle \mathcal{S}, \mathcal{A}, \mathcal{P}, \mathcal{R}, \gamma \rangle$. At each time step $t$, the agent observes a state $s_t \in \mathcal{S}$, selects an action $a_t \in \mathcal{A}$, receives a step-level reward $r_t = \mathcal{R}(s_t, a_t)$, and the environment transitions to a new state $s_{t+1} \sim \mathcal{P}(\cdot | s_t, a_t)$. The objective is to learn a policy $\pi_\theta$ parameterized by $\theta$ that maximizes the expected cumulative discounted return $J(\theta) = \mathbb{E}_{\tau \sim \pi_\theta} [\sum_{t=0}^\infty \gamma^t r_t]$. 

Unlike traditional reinforcement learning agents ~\citep{alphago,alphazero} that directly map states to actions, LLM agents possess the capability to perform explicit reasoning before acting. We formulate the policy of LLM agents as a joint distribution over a reasoning chain $z_t$ and an action $a_t$:
\begin{equation}
    \pi_\theta(z_t, a_t | s_t) = \pi_\theta(z_t | s_t) \cdot \pi_\theta(a_t | s_t, z_t),
\end{equation}
where $z_t$ represents a sequence of intermediate reasoning tokens (e.g., state analysis, future prediction, sub-goal decomposition) generated by the LLM. We represent the all output tokens as $c_t$, i.e., $c_t = \langle z_t, a_t \rangle$. This formulation decomposes the decision-making process into two stages: \textit{deliberation} ($\pi_\theta(z|s)$) and \textit{execution} ($\pi_\theta(a|s,z)$).

\paragraph{Our Approach}
In long-horizon interactive environments, optimal decision-making often requires looking ahead multiple steps into the future. An LLM agent attempts to emulate this by generating a reasoning chain $z_t$ that implicitly models the environment's transition dynamics. However, a critical discrepancy arises between the agent's internal world model and the actual environment dynamics. If the agent generates a reasoning chain based on a hallucinated transition $P_{model}(s'|s, a) \neq \mathcal{P}(s'|s, a)$, the resulting plan becomes invalid. As the lookahead depth increases, these errors compound exponentially, a phenomenon we term \textit{simulation drift}. To address this, we propose ProAct, a framework that grounds the  internal reasoning of agent in external reality, as shown in Figure~\ref{fig:proact_overview}. We aim to optimize $\pi_\theta$ such that the generated reasoning $z_t$ accurately reflects the true future value of actions without requiring expensive searching at inference time. This is achieved through two phases:
\begin{enumerate}
    \item \textbf{Grounded Lookahead Distillation:} We supervise the deliberation policy $\pi_\theta(z|s)$ using compressed trajectories derived from a search algorithm (e.g., MCTS) interacting with the environment $\mathcal{P}$. 
    \item \textbf{Monte-Carlo Critic:} We further refine the agent using online reinforcement learning, where the value estimation is calibrated by a Monte-Carlo Critic that offers low-variance advantage estimation, ensuring stable RL training.
\end{enumerate}

\subsection{Grounded Lookahead Distillation}
\label{sec:glad}

The intuition of GLAD is that long-horizon decision-making of interactive environments relies on looking ahead, simulating potential futures before committing to an action. However, LLM agents suffer from simulation drift, where their internal world models diverge from reality over long horizons. 
Instead of relying on the internal hallucinations of agents, we propose to externalize the lookahead process during reasoning data generation. We construct reasoning trajectories by allowing the model to read real future outcomes provided by the environment, and then compress this explicit search into an implicit intuition for future trends.

\subsubsection{Constructing Reasoning via Environmental Probing}
To generate high-quality supervision data, we design an iterative interaction loop where the environment acts as an oracle to ground the reasoning process of the model. This process consists of $k$ rounds of deliberation for a single decision:

\paragraph{Step 1: Environment-Augmented Lookahead.}
At each reasoning step $t$, rather than letting the model guess the future, we explicitly probe the real environment. We execute a Monte-Carlo Tree Search (MCTS) starting from the current state $s_t$. We sample $N$ trajectories $\{\tau_1, ..., \tau_N\}$, where each trajectory $\tau_i = (s_t, a_t^i, r_t^i, s_{t+1}^i, ...)$ extends for $T$ steps. 
Crucially, we record both optimal and suboptimal/dead-end paths. These trajectories serve as a ``ground-truth future map."

\paragraph{Step 2: Trajectory-Aware Decision Making.}
We feed the current state $s_t$ and the sampled raw trajectories directly into the context of LLM agent. The model is prompted to analyze and simulate these futures (e.g., ``Trajectory A leads to a merge, while Trajectory B leads to a gridlock").
Based on this unbiased environmental feedback, the model outputs:
\begin{enumerate}
    \item \textbf{Analysis:} A comparison and simulation of the potential futures.
    \item \textbf{Decision:} The next action to take, or a special token \texttt{<BACKTRACK>} if the analysis reveals that the current branch is suboptimal compared to previous alternatives.
\end{enumerate}
This ``Probing-Decision-Reflection" loop allows the model to correct its course if it identifies a bad state, effectively simulating a deep, self-correcting thought process grounded in reality.

\subsubsection{Cognitive Compression}
The raw interaction contexts from Phase 1 contain verbose search traces, structural tags, and backtracking steps. Directly fine-tuning on this data is inefficient and prone to overfitting. We introduce a compression step to distill the explicit search into a concise reasoning chain that estimates future trends.

We employ a teacher model (or the model itself) to synthesize the raw contexts into a final reasoning path $z$, adhering to four strict principles:

\begin{enumerate}
    \item \textbf{Format Simplification:} We remove all structural artifacts. The reasoning is rewritten in natural language (e.g., ``Let's analyze the board...", ``If I move up, the tiles will merge...") to align with the pre-trained distribution of LLM.
    
    \item \textbf{Explicit Reasoning Chains:} Each step must follow a strict causal logic: \textit{Observation} $\to$ \textit{Analysis} $\to$ \textit{Conclusion}. The analysis must explicitly link current actions to future states based on the  rules of environment observed in Phase 1.
    
    \item \textbf{Future Trend Estimation:} The compressed reasoning must not only explain the chosen action but also explain why other actions were rejected. For example, ``Moving left is safe now but blocks a critical merge in the future." This forces the model to internalize the environmental dynamics and learn counterfactual reasoning.
    
    \item \textbf{Preserve Diversity:} The reasoning $z$ must retain the ``trade-off" analysis found in the search. Instead of dogmatically stating the answer, it should reflect the deliberation process (e.g., ``Option A is good for score, but Option B provides better safety. Considering the long term, I choose B."), preserving the diversity of thought.
\end{enumerate}

Finally, we perform Supervised Fine-Tuning (SFT) on the dataset $\mathcal{D} = \{(s, z_{compressed}, a)\}$, minimizing the standard negative log-likelihood loss. This process effectively teaches the model to hallucinate correct future trends without needing to perform expensive search at inference time.

\begin{center} 
    \begin{minipage}{1\linewidth} 
        \begin{algorithm}[H] 
            \caption{ProAct: Grounded Lookahead Distillation}
            \label{alg:glad}
            \begin{algorithmic}[1]
            \State \textbf{Input:} Environment $\mathcal{E}$, Agent Policy $\pi_\theta$, Random Sampler
            \State \textbf{Output:} Dataset Buffer $\mathcal{D} \leftarrow \emptyset$
            
            \For{episode $e = 1$ to $M$}
                \State Initialize state $s_0$, History $H \leftarrow \emptyset$
                \State $t \leftarrow 0$ 
                \While{not $Done$}
                    \State $\mathcal{T}_{\text{real}} \leftarrow \text{ProbeEnvironment}(s_t, \text{depth}=d, \text{samples}=k)$ 
                    \State $\mathit{raw\_analysis}, a_t \leftarrow \pi_\theta(\text{Prompt}(s_t, \mathcal{T}_{\text{real}}))$
                    \If{$a_t == \texttt{<BACKTRACK>}$}
                        \State Revert $s_t$ to previous state $s_{t-1}$ (if applicable)
                        \State Continue to next iteration
                    \Else
                        \State Execute action: $s_{t+1}, r_t \leftarrow \mathcal{E}.\text{step}(a_t)$
                        \State Store interaction: $H.\text{append}((s_t, \mathcal{T}_{\text{real}}, \mathit{raw\_analysis}, a_t))$
                        \State $t \leftarrow t + 1$
                    \EndIf
                \EndWhile
                \For{each step $(s_t, \mathcal{T}_{\text{real}}, \mathit{raw\_analysis}, a_t)$ in $H$}
                    \State $z_{\text{compressed}} \leftarrow \text{Compress}(s_t, \mathcal{T}_{\text{real}}, a_t)$ 
                    \State Add to dataset: $\mathcal{D} \leftarrow \mathcal{D} \cup \{(s_t, z_{\text{compressed}}, a_t)\}$
                \EndFor
            \EndFor
            
            \end{algorithmic}
        \end{algorithm}
    \end{minipage}
\end{center}

\subsection{Online Reinforcement Learning with Monte-Carlo Critic}
\label{sec:mc-critic}

Reinforcement Learning (RL) has recently become a prominent method for empowering LLM agents ~\citep{deepseek-r1,openai-o1}. However, despite its efficacy, applying RL to LLMs presents significant hurdles, particularly regarding value function (i.e., critic network) estimation in long-horizon scenarios. Formally, the state-value function $V_{\pi}(s_t)$ represents the expected cumulative reward an agent obtains starting from state $s_t$ and following policy $\pi$. Similarly, the action-value function $Q_{\pi}(s_t, a_t)$ denotes the expected cumulative reward starting from state $s_t$, taking action $a_t$, and thereafter following policy $\pi$. These two value functions are defined as follows:
\begin{align}
    V_{\pi}(s_t) &=  \mathbb{E}_{\tau \sim \pi} \left[ \sum_{k=t}^\infty \gamma^{k-t} r_k \right], \\
    Q_{\pi}(s_t, a_t) &= \mathbb{E}_{s_{t+1} \sim \mathcal{P}(\cdot | s_t, a_t)} \left[ r_t + \gamma V_{\pi}(s_{t+1})\right]. \label{eq:q_function}
\end{align}
In traditional deep RL, policy networks typically consist of simple Multi-Layer Perceptrons (MLPs) with millions of parameters. This lightweight architecture permits rapid environment interaction, allowing agents to quickly collect millions of steps of interaction data for training. This high sample throughput is crucial for training a critic network that produces precise value estimates. As a result, traditional deep RL has demonstrated superhuman performance across various domains ~\citep{alphago,alphastar,openaifive,wzry}. In contrast, LLMs contain parameters on the scale of billions, and rollouting a single step often entails the autoregressive synthesis of hundreds or thousands of tokens. Consequently, the speed of online interaction and sample generation is prohibitively slow. This limitation leads to high variance in the value estimates produced by the critic network, frequently resulting in training instability. To address this challenge, we propose a Monte-Carlo Critic (MC-Critic) designed to acquire low-variance value estimates in a rapid and cost-effective way.

\subsubsection{MC-Critic Formulation}

Instead of training a parameterized critic network to approximate the value $V_{\pi_\theta}(s_t)$, MC-Critic is a parameter-free critic that estimates value directly via Monte Carlo rollouts. Specifically, given a state $s_t$, the LLM agent interacts with the environment following the current policy $\pi_\theta$ to generate $M$ trajectories $\{\tau_1, \cdots, \tau_M\}$, where each trajectory $\tau_i = \{s_t, c_t^i, r_t^i, s_{t+1}^i, c_{t+1}^i, r_{t+1}^i, \cdots, s_{t+T}^i\}$ extends for a maximum of $T$ time steps. MC-Critic then calculates the value of state $s_t$ as the average of the cumulative discounted rewards over these $M$ trajectories, denoted as:
\begin{align}
    V_{\pi_{\theta}}^{\text{MC}}(s_t) = \frac{1}{M} \sum_{i=1}^M \sum_{k=0}^{T-1} \gamma^{k} r_{k+t}^i. \label{eq:mc_critic}
\end{align}
This approach yields an unbiased value estimator, where variance can be reduced by increasing $M$. However, our empirical tests indicate that a 4B parameter LLM requires 3–6 seconds to generate a single-step reasoning chain and action. Consequently, utilizing the LLM to rollout $M$ trajectories for estimating the value of a single state is prohibitively expensive and time-consuming. To mitigate this inference latency, we introduce a lightweight Monte Carlo rollout scheme that employs a random policy $\pi_{random}$ as a surrogate for the LLM policy $\pi_\theta$ to generate the $M$ trajectories. We then derive the value estimate $V_{\pi_{random}}^{\text{MC}}(s_t)$ following Eq.~\ref{eq:mc_critic}. Although $V_{\pi_{random}}^{\text{MC}}(s_t)$ is theoretically suboptimal compared to $V_{\pi_{\theta}}^{\text{MC}}(s_t)$, it offers a low-variance value estimate very quickly and efficiently. For instance, in the game 2048, the random policy can rollout over 1,000 trajectories in less than 3 seconds.

\subsubsection{Multi-Turn RL Optimization}

MC-Critic serves as a plug-and-play auxiliary value estimator, which is adaptable to any reinforcement learning algorithm requiring state value estimation. Below, we detail the integration of MC-Critic with two prevalent RL algorithms for LLMs: GRPO ~\citep{grpo} and PPO ~\citep{ppo}, referring to the resulting variants as MC-GRPO and MC-PPO.

\paragraph{MC-GRPO}

The most straightforward way to extend GRPO to multi-turn scenarios is Trajectory-Level GRPO ~\citep{ragen} (hereinafter referred to as Traj-GRPO). In Traj-GRPO, $G$ trajectories $\{\tau_i\}_{i=1}^G$ are generated starting from a shared initial state $s_0$. For each trajectory $\tau_i=\{s_0, c_0^i, r_0^i, s_1^i, \cdots, s_{T_i}^i\}$, a trajectory-level reward $R(\tau_i)=\sum_{t=0}^{T_i-1} r_t^i$ is computed, where $T_i$ represents the number of time steps in the trajectory $\tau_i$. Subsequently, group normalization is applied to these trajectory-level rewards to derive trajectory-level advantages:
\begin{align}
    \hat{A}^{\text{Traj-GRPO}}_i = \frac{R(\tau_i) - \text{mean}(\{R(\tau_1), \cdots, R(\tau_G)\})}{\text{std}(\{R(\tau_1), \cdots, R(\tau_G)\})}.
\end{align}
To simplify the loss formulation, let us define the probability ratio $\rho_{t,j}^i(\theta)$ as:
\begin{align}
\rho_{t,j}^i(\theta) = \frac{\pi_\theta(c_{t,j}^i|s_t^i, c_{t,<j}^i)}{\pi_{\theta_{\text{old}}}(c_{t,j}^i|s_t^i, c_{t,<j}^i)}, \label{eq:prob_ratio}
\end{align}
where $c_{t,j}^i$ is the $j^{\text{th}}$ token of $c_{t}^i$ and $c_{t,<j}^i$ are the tokens before $c_{t,j}^i$. The loss function for Traj-GRPO is then defined as:
\begin{small}
\begin{equation}
\begin{aligned}
    \mathcal{L}^{\text{Traj-GRPO}}(\theta) = - \frac{1}{\sum_{i=1}^G \sum_{t=0}^{T_i-1} |c_{t}^i|} \sum_{i=1}^G \sum_{t=0}^{T_i-1}  \sum_{j=1}^{|c_{t}^i|} \min \left[ \rho_{t,j}^i(\theta) \hat{A}^{\text{Traj-GRPO}}_i,  \text{clip}\left( \rho_{t,j}^i(\theta), 1-\epsilon, 1+\epsilon  \right)\hat{A}^{\text{Traj-GRPO}}_i \right],
\end{aligned}
\end{equation}
\end{small}%
which simply assigns the shared trajectory-level advantage $\hat{A}^{\text{Traj-GRPO}}_i$ to all time steps outputs $c_{t}^i$ within trajectory $\tau_i$, lacking discriminative credit assignment for different time steps. This can easily lead to model collapse ~\citep{turn-ppo}. Furthermore, it utilizes the trajectory-level reward and does not involve estimating the value of individual states, making it difficult to combine with MC-Critic. 

To this end, we consider a step-level variant of GRPO for multi-turn scenarios, termed Step-GRPO. It rollouts a group of single-step samples instead of a group of trajectories. Specifically, we first rollout a trajectory and store all visited states in a state pool, which functions similarly to the question set in the math domain ~\citep{grpo}. During each training step, we randomly select a batch of states $\{s_{t_u}\}_{u=1}^b$ from the state pool. For each state $s_{t_u}$, we generate $G$ independent single-step samples $\{s_{t_u}, c_{t_u}^i, r_{t_u}^i\}_{i=1}^G$. Then, we compute the advantages for the single-step samples based on the step-level rewards $\{r_{t_u}^i\}_{i=1}^G$:
\begin{align}
    \hat{A}^{\text{Step-GRPO}}_{t_u,i} = \frac{r_{t_u}^i - \text{mean}(\{r_{t_u}^1, \cdots, r_{t_u}^G\})}{\text{std}(\{r_{t_u}^1, \cdots, r_{t_u}^G\})}.
\end{align}
The loss function for Step-GRPO is as follows:
\begin{small}
\begin{equation}
\begin{aligned}
    \mathcal{L}^{\text{Step-GRPO}}(\theta) = - \frac{1}{\sum_{u=1}^{b} \sum_{i=1}^G |c_{t_u}^i|} \sum_{u=1}^{b} \sum_{i=1}^G  \sum_{j=1}^{|c_{t_u}^i|} \min \left[ \rho_{t_u,j}^i(\theta) \hat{A}^{\text{Step-GRPO}}_{t_u,i}, \text{clip}\left( \rho_{t_u,j}^i(\theta), 1-\epsilon, 1+\epsilon  \right)\hat{A}^{\text{Step-GRPO}}_{t_u,i} \right].
\end{aligned}
\end{equation}
\end{small}%
Building upon Step-GRPO, we introduce MC-GRPO, which augments Step-GRPO with the MC-Critic. The primary distinction lies in the substitution of the immediate step-reward with an action-value function to compute the advantage for the single-step sample $\{s_{t_u}, c_{t_u}^i, r_{t_u}^i\}$. Specifically, the action-value function for $\{s_{t_u}, c_{t_u}^i, r_{t_u}^i\}$ (i.e., $\{s_{t_u}, z_{t_u}^i, a_{t_u}^i, r_{t_u}^i\}$) is estimated via the MC-Critic as follows: 
\begin{align}
    Q_{\pi_{random}}^{\text{MC}}(s_{t_u}, a_{t_u}^i) &= \mathbb{E}_{s_{t_u+1} \sim \mathcal{P}(\cdot | s_{t_u}, a_{t_u}^i)} \left[ r_{t_u}^i + \gamma  V_{\pi_{random}}^{\text{MC}}(s_{t_u+1}) \right].
    \label{eq:q_mc}
\end{align}
Subsequently, we apply group normalization to derive the advantage: 
\begin{align}
    \hat{A}^{\text{MC-GRPO-relative}}_{t_u,i} = \frac{Q_{\pi_{random}}^{\text{MC}}(s_{t_u}, a_{t_u}^i) - \text{mean}(\{Q_{\pi_{random}}^{\text{MC}}(s_{t_u}, a_{t_u}^i)\}_{i=1}^G)}{\text{std}(\{Q_{\pi_{random}}^{\text{MC}}(s_{t_u}, a_{t_u}^i)\}_{i=1}^G)}.
\end{align}
In practice, we found that if the actions $\{a_{t_u}^i\}_{i=1}^G$ selected within a group are identical, their corresponding action-values are also identical. This results in an advantage of zero for the entire group (i.e., zero policy gradients), making it difficult to improve the policy for that state $s_{t_u}$. Instead of the common practice of discarding these samples through dynamic sampling ~\citep{dapo}, we aim to leverage these group samples to further enhance the policy. Specifically, when $\{a_{t_u}^i\}_{i=1}^G$ are identical, we replace the group-relative baseline with the average of action-values of all actions in the action space $\mathcal{A}$ (i.e., an absolute baseline) to compute the advantage:
\begin{align}
    \hat{A}^{\text{MC-GRPO-absolute}}_{t_u,i} = \frac{Q_{\pi_{random}}^{\text{MC}}(s_{t_u}, a_{t_u}^i) - \text{mean}(\{Q_{\pi_{random}}^{\text{MC}}(s_{t_u}, a)\}_{a \in \mathcal{A}})}{\text{std}(\{Q_{\pi_{random}}^{\text{MC}}(s_{t_u}, a)\}_{a \in \mathcal{A}})}.
\end{align}
The final advantage formulation for MC-GRPO is defined as:
$$\hat{A}^{\text{MC-GRPO}}_{t_u,i} =
\begin{cases}
\hat{A}^{\text{MC-GRPO-relative}}_{t_u,i} & \text{if $\{a_{t_u}^i\}_{i=1}^G$ are not identical,} \\
\hat{A}^{\text{MC-GRPO-absolute}}_{t_u,i} & \text{if $\{a_{t_u}^i\}_{i=1}^G$ are identical.}
\end{cases}$$
This yields the loss function for MC-GRPO: 
\begin{small}
\begin{equation}
\begin{aligned}
    \mathcal{L}^{\text{MC-GRPO}}(\theta) = - \frac{1}{\sum_{u=1}^{b} \sum_{i=1}^G |c_{t_u}^i|} \sum_{u=1}^{b} \sum_{i=1}^G  \sum_{j=1}^{|c_{t_u}^i|} \min \left[ \rho_{t_u,j}^i(\theta) \hat{A}^{\text{MC-GRPO}}_{t_u,i}, \text{clip}\left( \rho_{t_u,j}^i(\theta), 1-\epsilon, 1+\epsilon  \right)\hat{A}^{\text{MC-GRPO}}_{t_u,i} \right].
\end{aligned}
\end{equation}
\end{small}%
The only difference between $\mathcal{L}^{\text{MC-GRPO}}$ and $\mathcal{L}^{\text{Step-GRPO}}$ is in the calculation of the advantage. By incorporating the MC-Critic, the LLM agent is incentivized to prioritize long-term returns over immediate single-step gains, thereby fostering multi-turn lookahead reasoning capabilities.

\paragraph{MC-PPO}

Recent advancements have extended PPO to multi-turn scenarios, such as VAGEN ~\citep{vagen} and Turn-PPO ~\citep{turn-ppo}. These methods typically concatenate the entire interaction history as context before the agent makes a decision at each time step. However, in multi-turn scenarios where a single trajectory contains hundreds of time steps, this practice often causes the context length to exceed the maximum token limit (e.g., 32,768 tokens). To address this, we adopt a simplified context formulation where the LLM agent relies solely on the current state for decision-making, discarding historical interaction information. Building upon this setting, we propose Step-PPO, a PPO variant tailored for multi-turn scenarios. 
Similar to Turn-PPO ~\citep{turn-ppo}, Step-PPO employs a turn-level critic $V_\phi$ to compute the turn-level GAE ~\citep{gae} advantage for policy updates, where all tokens within a turn share the same turn-level advantage. Specifically, we first rollout a trajectory $\tau=(s_0, c_0, r_0, s_1, c_1, r_1, \dots, s_T)$ and designate the critic's output at the final token of the response $c_t$ as the turn-level value $V_\phi(s_t, c_t)$. We then calculate the advantage for each turn using GAE:
\begin{equation}
    \delta_{t}^{\text{Step-PPO}} = r_{t} + \gamma V_\phi(s_{t+1}, c_{t+1}) - V_\phi(s_t, c_{t}),
\end{equation}
\begin{equation}
    \hat{A}_{t}^{\text{Step-PPO}} = \sum_{l=t}^{T-1} (\gamma\lambda)^{l-t} \delta_{l}^{\text{Step-PPO}},
\end{equation}
Subsequently, the policy loss for Step-PPO is defined as:
\begin{small}
\begin{equation}
\begin{aligned}
    \mathcal{L}^{\text{Step-PPO}}(\theta) = - \frac{1}{\sum_{t=0}^{T-1} |c_{t}|} \sum_{t=0}^{T-1}  \sum_{j=1}^{|c_{t}|} \min \left[ \rho_{t,j}(\theta) \hat{A}_{t}^{\text{Step-PPO}},  \text{clip}\left( \rho_{t,j}(\theta), 1-\epsilon, 1+\epsilon  \right)\hat{A}_{t}^{\text{Step-PPO}} \right],
\end{aligned}
\end{equation}
\end{small}%
and the value loss is given by:
\begin{small}
\begin{equation}
\begin{aligned}
    \mathcal{L}^{\text{Step-PPO}}(\phi) = \frac{1}{T} \sum_{t=0}^{T-1}  \left( V_\phi(s_t, c_t) - R_t \right)^2, \,\,\, \text{where} \,\,\, R_t = \hat{A}_{t}^{\text{Step-PPO}} + V_{\phi_{\text{old}}}(s_t, c_t).
\end{aligned}
\end{equation}
\end{small}%
As previously discussed, the inherent sample inefficiency of LLMs results in high variance in the values estimated by the trained critic $V_\phi$. To mitigate this, we propose MC-PPO, which leverages the MC-Critic to assist in estimating a turn-level value with reduced variance. Concretely, we compute a weighted sum of the value $V_{\pi_{random}}^{\text{MC}}(s_t)$ estimated by the MC-Critic and the value $V_\phi(s_t, c_t)$ output by the turn-level critic to derive the step-level value for MC-PPO:
\begin{equation}
    V^{\text{MC-PPO}}(s_t) = (1 - \omega) V_\phi(s_t, c_t) + \omega V_{\pi_{random}}^{\text{MC}}(s_t),
\end{equation}
where $\omega \in [0, 1]$ represents the weight of the value estimated by MC-Critic. We then use this new value to calculate the advantage and update the policy and the critic, identical to the process in Step-PPO.

\section{Experiments}
We evaluate ProAct on two representative long-horizon decision-making benchmarks: \textit{2048}, a stochastic environment that requires planning under uncertainty and contains hundreds of turns in a single trajectory, and \textit{Sokoban}, a deterministic planning task with shorter turns per trajectory but sparse rewards. These environments pose complementary challenges for lookahead reasoning and error accumulation. To evaluate generalization beyond the training distribution, we further test ProAct on multiple environment variants.

\subsection{Experimental Setup}

\subsubsection{Environments}
The LLM agent interacts with both environments through textual observations and action descriptions. Environment states are serialized into structured text following a consistent format across all methods. Concrete examples are provided in Appendix~\ref{apend:env}.

\paragraph{2048.}
A single-agent puzzle environment defined on a $4 \times 4$ grid. At each step, the agent selects one of four actions: \textit{up}, \textit{down}, \textit{left}, or \textit{right}. After each move, a new tile is randomly spawned on an empty cell, introducing stochasticity into the environment. Tiles with identical values merge upon contact, producing a tile with their summed value. 
The step-level reward is set as follows: if the output format is incorrect, the agent receives a penalty of -10. If the output format is correct but the parsed action is invalid (i.e., no tiles are moved), a penalty of -1 is given. Otherwise, the reward is equal to the sum of values of all new merged tiles. The maximum time steps of a trajectory is set to $1000$. A trajectory will terminate when the maximum time steps is exceeded or when 10 consecutive invalid actions are performed.
The objective is to maximize the value of the highest tile obtained during a trajectory.

We consider several environment variants to evaluate generalization, including changing the grid size from $4 \times 4$ to $3 \times 3$, and modifying the minimum tile value from 2 to 3, which we refer to as \textit{3072}.

\paragraph{Sokoban.} A classical deterministic puzzle environment in which the agent must push all boxes onto designated target locations. 
The agent can move between empty cells using four directional actions: \textit{up}, \textit{down}, \textit{left}, and \textit{right}. 
When adjacent to a box, the agent can execute a \textit{push} action in a given direction, where \textit{push} is prefixed to the directional command (e.g., \textit{push up}, \textit{push down}, \textit{push left}, \textit{push right}), provided that the target cell is empty. The maximum time steps of a trajectory is set to $200$.
A trajectory terminates successfully when all boxes are placed on their target positions.
For the step-level reward, the agent receives a penalty of $-2$ if it produces an invalid output, including failing to follow the required output format or generating an invalid action. Otherwise, it obtains a reward of $+1$ for each box pushed onto a target location, a penalty of $-1$ for pushing a box off a target, and a terminal reward of $+10$ for successfully completing the level.

To evaluate generalization, we consider three types of Sokoban variants: 
(1) unseen levels that are not encountered during training, 
(2) action space modifications where boxes are pushed directly using directional actions, and 
(3) change the symbolic representation of the Sokoban map.

\subsubsection{Model}
All experiments are conducted using \textit{Qwen3-4B-Instruct} ~\citep{qwen3} as the backbone language model. Unless otherwise specified, all baselines and ablations share the same model architecture and parameter count to ensure a fair comparison. The model is fine-tuned end-to-end without freezing any parameters.

\subsubsection{Training Protocol}
The training process of ProAct consists of two stages: SFT with GLAD, followed by RL enhanced with MC-Critic.

\paragraph{GLAD.}
In the first stage, we construct a supervised training dataset by generating trajectories using search-based agents. To promote diversity, 2048 trajectories are initialized with different random seeds, while Sokoban levels are procedurally generated. Each trajectory step is stored as a training sample containing the textual environment state, the distilled lookahead reasoning chain, and the resulting action. In total, we collect 25K samples for 2048 and 8K samples for Sokoban to form the SFT dataset.


\paragraph{MC-Critic.}
While GLAD equips the model with grounded lookahead priors, it does not directly optimize long-horizon returns. We therefore introduce MC-Critic to further refine decision-making via RL. MC-Critic is integrated with two standard RL algorithms, PPO and GRPO. 


We conduct two groups of RL training with different initializations. In the first setting, the policy is initialized from the SFT checkpoint and further fine-tuned using RL on the same Sokoban levels (Base) used for SFT evaluation. To assess generalization, we additionally evaluate the resulting policy on a disjoint set of unseen Sokoban levels (Extra).
In the second setting, training starts from the original Qwen3-4B-Instruct weights without SFT initialization. 
However, when training from the original Qwen3-4B-Instruct weights without SFT initialization, we find that Sokoban levels aligned with the SFT training and evaluation difficulty are overly challenging under this initialization; therefore, we adopt a set of simplified Sokoban levels that can be solved within 20 steps, and accordingly set the maximum time steps of a trajectory to $20$.
All SFT and RL experiments are implemented within the AReaL framework ~\citep{areal}. Detailed training configurations are provided in Appendix~\ref{apend:train_config}.

\subsubsection{Evaluation}
We evaluate agent performance using environment-specific scoring metrics. In 2048, performance is measured by the cumulative score obtained from tile merge operations, where each merge contributes the value of the resulting tile. In Sokoban, performance is measured by the average number of boxes successfully pushed onto target locations per level. We use average boxes placed rather than binary success to provide a smoother signal that reflects partial progress on harder levels. For both environments, each setting is evaluated over multiple runs and the mean score is reported. More details are provided in Appendix~\ref{apend:env}.

We compare ProAct against a range of model baselines, including both closed-source and open-source Instruct Models. The full list of evaluated models is reported in Table~\ref{tab:sft_main}. In addition, we evaluate MC-Critic against standard RL algorithms, such as PPO and GRPO, to verify its effectiveness. All models are evaluated under the same environment interface, with the temperature fixed to 0.6.

For Sokoban, we use a fixed set of levels that do not appear in the SFT training data as the base evaluation benchmark. To maintain consistency between evaluation and reinforcement learning, these levels are also used as the training environments in the subsequent RL phase.

\subsection{Results}

\subsubsection{GLAD}

\paragraph{Supervised Training.}
As shown in Table~\ref{tab:sft_main}, our 4B GLAD-trained model consistently outperforms all open-source baselines and several strong closed-source models. 
Importantly, GLAD exhibits robustness to environment variations: for 2048, the model trained on the standard 4$\times$4 grid also achieves consistent gains on reduced 3$\times$3 grids and on the 3072 variant; 
for Sokoban, the model generalizes effectively both to the unseen levels (Base) and to their variants with modified action spaces and symbolic representations. These results indicate that the distilled lookahead reasoning learned via GLAD enables substantial margin over baselines both in-distribution and under challenging out-of-distribution scenarios.

\begin{table*}[t]
\centering
\caption{Performance comparison on 2048 and Sokoban under multiple environment variants.
For 2048, \textbf{4$\times$4} denotes the standard $4\times4$ grid setting, 
\textbf{3$\times$3} corresponds to a reduced grid-size variant, 
and \textbf{3072} indicates a modified environment where the minimum tile value is 3, 
resulting in a target tile of 3072 instead of 2048.
For Sokoban, \textbf{Base} evaluates performance on newly generated levels that do not appear in the SFT training data,
\textbf{Action} modifies the action space,
and \textbf{Symbol} alters the symbolic representation of the map. Columns marked with $^*$ indicate evaluation on environment variants that are not seen during SFT and RL training.}
\label{tab:sft_main}
\small
\setlength{\tabcolsep}{5pt}
\begin{tabular}{lcccccc}
\toprule
 & \multicolumn{3}{c}{\textbf{2048}} & \multicolumn{3}{c}{\textbf{Sokoban}} \\
\cmidrule(lr){2-4} \cmidrule(lr){5-7}
\textbf{Model} 
& \textbf{4$\times$4} 
& \textbf{3$\times$3$^*$} 
& \textbf{3072$^*$}
& \textbf{Base} 
& \textbf{Action$^*$} 
& \textbf{Symbol$^*$} \\
\midrule
\multicolumn{7}{l}{\textbf{Closed-source Models}} \\
\midrule

GPT-5 \citep{gpt5} 
& 4040.0 & \textbf{1184.0} & \textbf{6962.0} 
& \textbf{1.89} & \textbf{1.83} & 1.94 \\

Claude-4.5-Sonnet \citep{claude4.5} 
& 166.7 & 109.3 & 120.0 
& 1.06 & 0.78 & 1.33 \\

Doubao-Seed-1.6 \citep{seed1.6} 
& 1877.3 & 300.0 & 2054.0 
& 1.22 & 0.61 & 1.56 \\

Doubao-Seed-1.8 \citep{seed1.8} 
& \textbf{4662.7} & 545.3 & 4210.0 
& 1.80 & 1.44 & \textbf{2.00} \\

UITARS-1.5 \citep{uitars1.5} 
& 2616.0 & 466.7 & 3920.0 
& 0.44 & 0.44 & 0.39 \\
\midrule

\multicolumn{7}{l}{\textbf{Open-source Models + Ours}} \\
\midrule

Qwen3-235B-A22B-Instruct 
& 634.7 & 274.7 & 2688.0 
& 0.61 & 0.33 & 0.56 \\

Qwen3-30B-A3B-Instruct 
& 838.7 & 230.7 & 1740.0 
& 0.44 & 0.39 & 0.50 \\

Qwen3-4B-Instruct (Base Model) 
& 721.3 & 187.3 & 1603.0 
& 0.39 & 0.44 & 0.56 \\

Base Model + GLAD
& 3335.3
& 429.2
& 4565.7
& 0.72
& 0.52
& 0.67\\

Base Model + GLAD + MC-Critic
& \textbf{4503.8}
& \textbf{464.4}
& \textbf{6013.7}
& \textbf{0.94}
& \textbf{0.62}
& \textbf{0.70}\\


\bottomrule
\end{tabular}
\end{table*}

\begin{figure}[h]
\centering
\includegraphics[width=1 \textwidth]{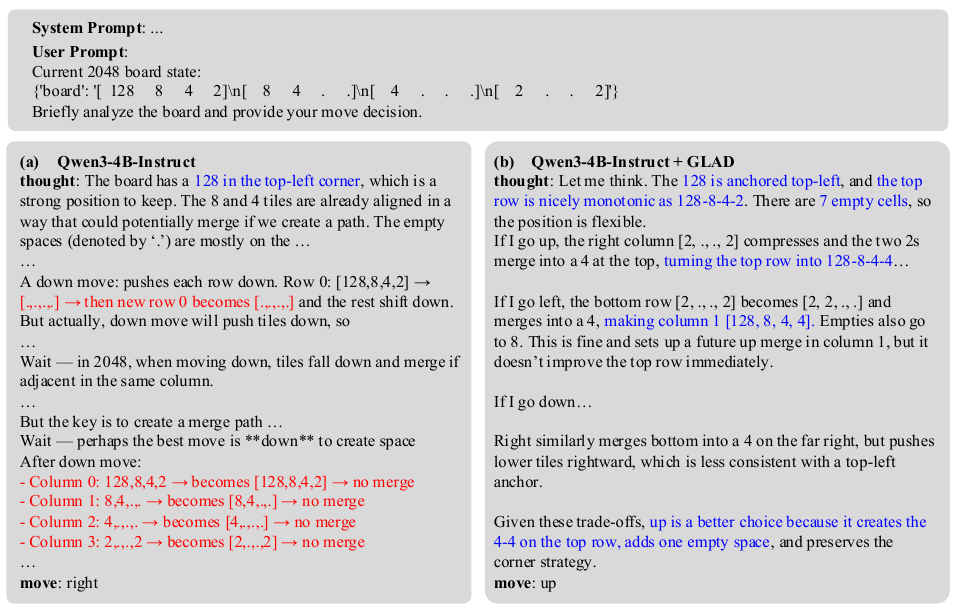}

\caption{
Case Study of GLAD on 2048. (a) Base model (Qwen3-4B-Instruct) and (b) GLAD-supervised model. \textcolor{blue}{Blue} and \textcolor{red}{red} segments denote correct and incorrect intermediate analysis, respectively.
}

\vspace{2em}
\label{fig:case_study}
\end{figure}

\paragraph{Case Study.} 
Figure~\ref{fig:case_study} illustrates a qualitative comparison of the reasoning behaviors of models before and after GLAD supervision on a representative 2048 task.
The model is prompted to generate its output in two stages: an intermediate analysis channel used to structure internal reasoning, followed by a final action prediction in the form \texttt{move: <action>}.

Figure~\ref{fig:case_study}(a) shows the behavior of the base model. Although it is an instruction-tuned model, its intermediate analysis exhibits characteristics commonly associated with generic “thinking” paradigms, including redundant waiting steps and unstable reasoning trajectories. Moreover, the base model occasionally hallucinates board configurations and transition outcomes, leading to incorrect or poorly justified action choices despite appearing verbose in its reasoning process. In contrast, Figure~\ref{fig:case_study}(b) presents the GLAD-supervised model. Its intermediate analysis is noticeably more compact and accurate. The model correctly describes the current board state in natural language, performs explicit lookahead by simulating the outcomes of multiple candidate actions, and systematically compares their consequences in terms of tile merging and future potential. Based on this comparison, the model selects the optimal action with a clear and consistent justification. The full prompt and response is given in Appendix \ref{apend:prompt}.

\subsubsection{MC-Critic}
We evaluate the effectiveness of MC-Critic in two complementary reinforcement learning settings:
(1) \emph{RL fine-tuning on top of GLAD-initialized models}, and
(2) \emph{training from scratch starting from the base instruction-tuned model}.
This design allows us to disentangle the contribution of MC-Critic from that of GLAD supervision and to assess whether MC-Critic alone can improve long-horizon optimization stability. 
Besides, we analyze two important hyperparameters used in MC-Critic and summarize the recipe of MC-Critic.

\paragraph{Train on GLAD-Initialized Models.}
We first investigate MC-Critic as an RL refinement applied on top of GLAD-initialized agents.
Starting from the GLAD SFT checkpoint, we fine-tune the model using PPO and GRPO, both with and without MC-Critic, in order to isolate its contribution.

\begin{figure}[t]
    \centering 
    
    \begin{subfigure}[b]{0.33\textwidth}
        \centering
        \includegraphics[width=\linewidth]{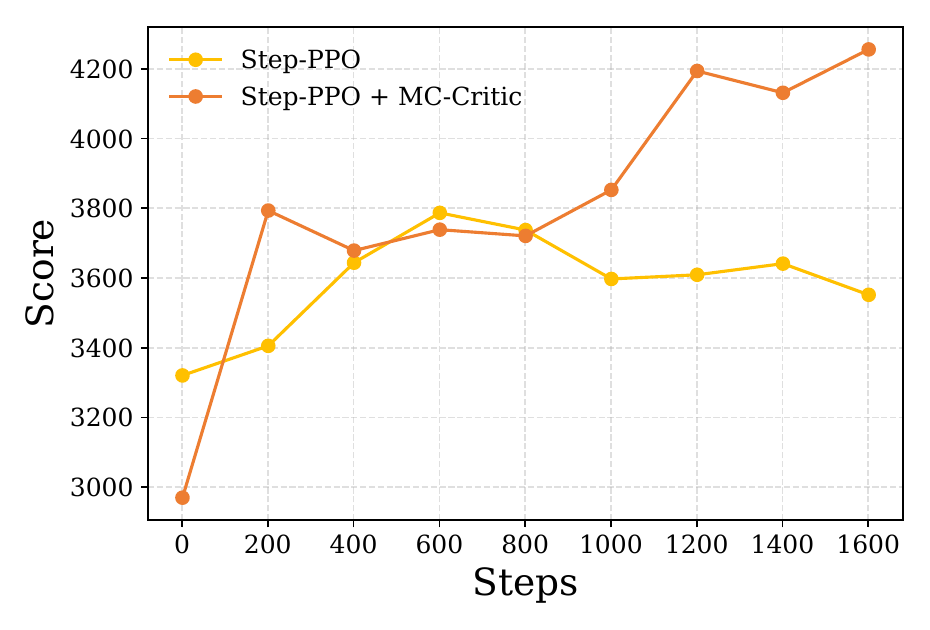}
        \caption{PPO on 2048}
        \label{fig:ppo_2048}
    \end{subfigure}
    \hspace{2em} 
    \begin{subfigure}[b]{0.33\textwidth}
        \centering
        \includegraphics[width=\linewidth]{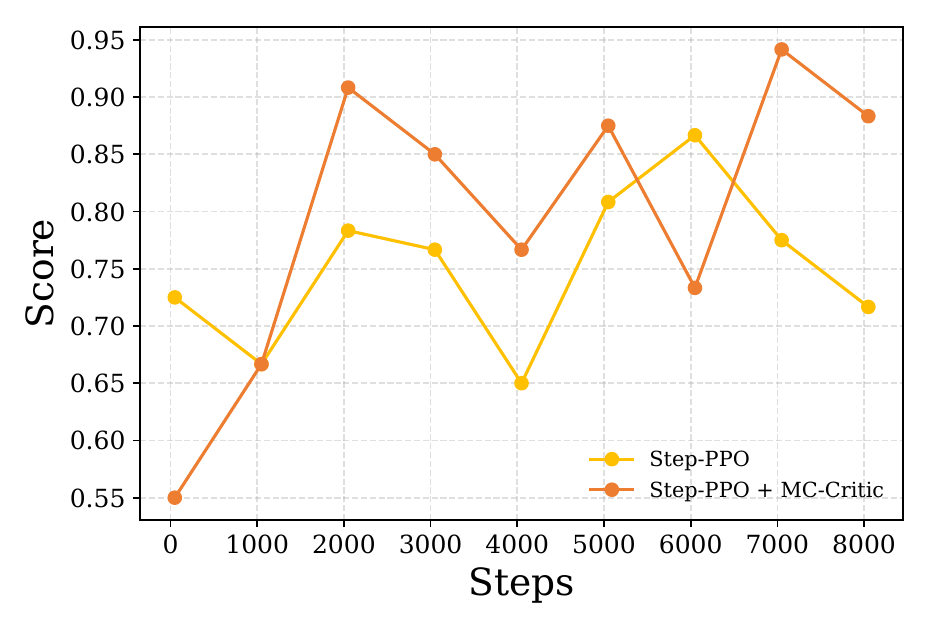}
        \caption{PPO on Sokoban}
        \label{fig:ppo_sokoban}
    \end{subfigure}
    
    \par\vspace{1em} 
    
    \begin{subfigure}[b]{0.33\textwidth}
        \centering
        \includegraphics[width=\linewidth]{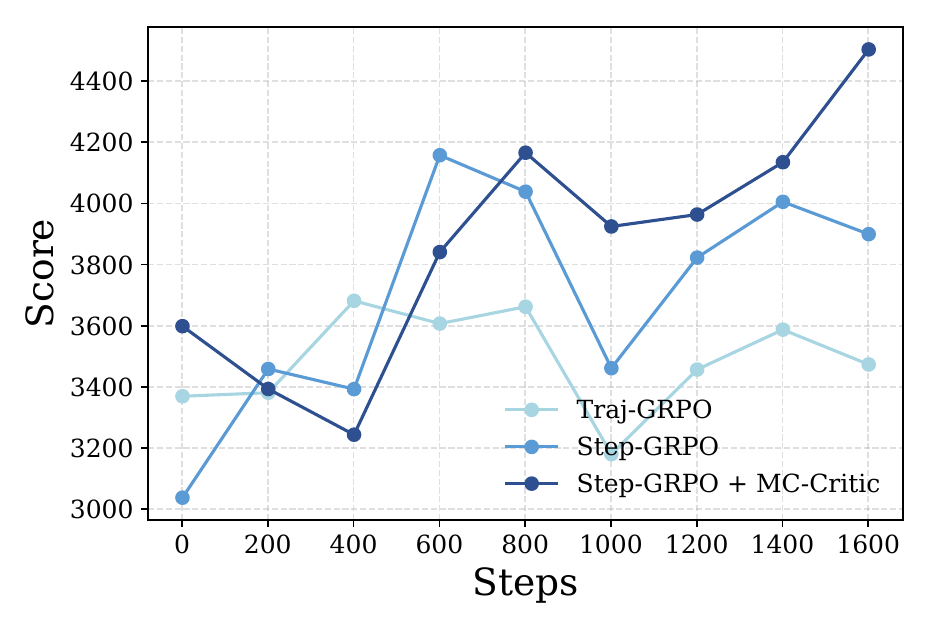}
        \caption{GRPO on 2048}
        \label{fig:grpo_2048}
    \end{subfigure}
    \hspace{2em} 
    \begin{subfigure}[b]{0.33\textwidth}
        \centering
        \includegraphics[width=\linewidth]{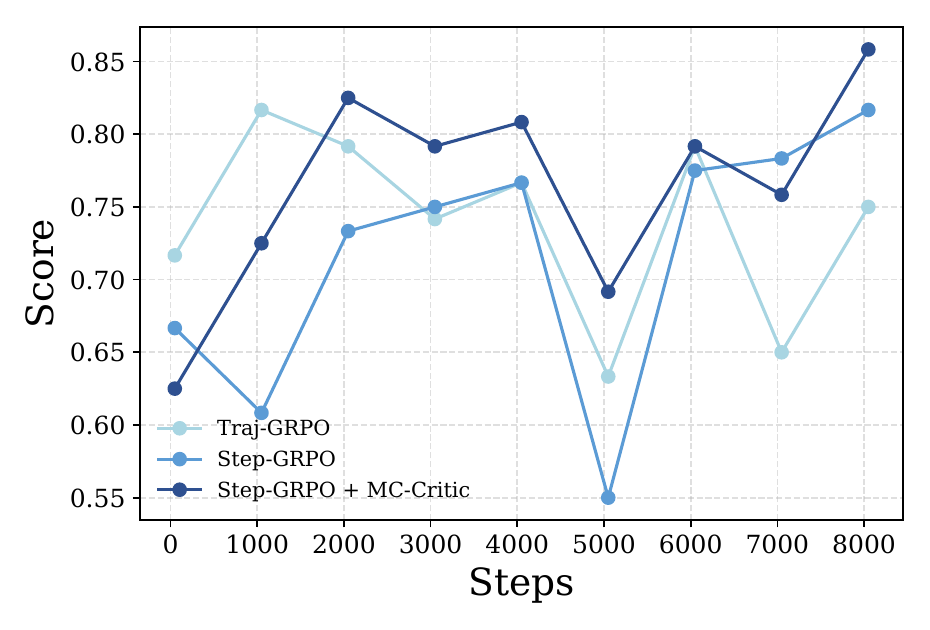}
        \caption{GRPO on Sokoban}
        \label{fig:grpo_sokoban}
    \end{subfigure}
    
    \caption{Comparison of baseline RL methods and MC-Critic when trained from GLAD SFT checkpoints on 2048 and Sokoban. MC-PPO and MC-GRPO are represented by ``Step-PPO + MC-Critic" and ``Step-GRPO + MC-Critic", respectively.}
    \label{fig:rl_sft}

\end{figure}

As shown in Figure~\ref{fig:rl_sft}, MC-Critic consistently improves performance on both 2048 and Sokoban.
Notably, the incorporation of MC-Critic benefits both GRPO and PPO, leading to higher final scores.
Moreover, as reported in Table~\ref{tab:variant_sft_results}, these improvements are not limited to the training distribution; performance gains consistently transfer to variant environments.

Overall, these results demonstrate that, when initialized from a strong GLAD prior, MC-Critic effectively exploits Monte-Carlo returns to further refine long-horizon decision-making, while maintaining robust generalization to out-of-distribution settings.




\begin{table}[t]
\centering
\footnotesize
\setlength{\tabcolsep}{4pt}
\begin{tabular}{lccccc}
\toprule
 & \multicolumn{2}{c}{2048} & \multicolumn{3}{c}{Sokoban} \\
\cmidrule(lr){2-3} \cmidrule(lr){4-6}
Method & 3$\times$3 & 3072 & Extra & Action & Symbol \\
\midrule
Traj-GRPO & 459.0 & 5457.0 & 1.10 & 0.60 & 0.64 \\
Step-GRPO & 457.0 & 5820.3 & 1.00 & 0.56 & 0.63 \\
Step-PPO  & \textbf{487.3} & 6248.7 & 0.90 & \textbf{0.62} & 0.65 \\
\midrule
MC-GRPO   & 464.4 & \textbf{6013.7} & 1.05 & 0.60 & \textbf{0.70} \\
MC-PPO    & 465.0 & 5818.1 & \textbf{1.18} & \textbf{0.62} & 0.68 \\
\bottomrule
\end{tabular}
\caption{Performance of RL methods with and without MC-Critic on 2048 and Sokoban environment variants when trained from scratch.}
\label{tab:variant_sft_results}
\end{table}

\paragraph{Train From Scratch.}
We next evaluate whether MC-Critic can enhance RL performance without GLAD supervision, training all methods directly from the base Qwen3-4B-Instruct model.

\begin{figure}[t]
    \centering 
    
    \begin{subfigure}[b]{0.33\textwidth}
        \centering
        \includegraphics[width=\linewidth]{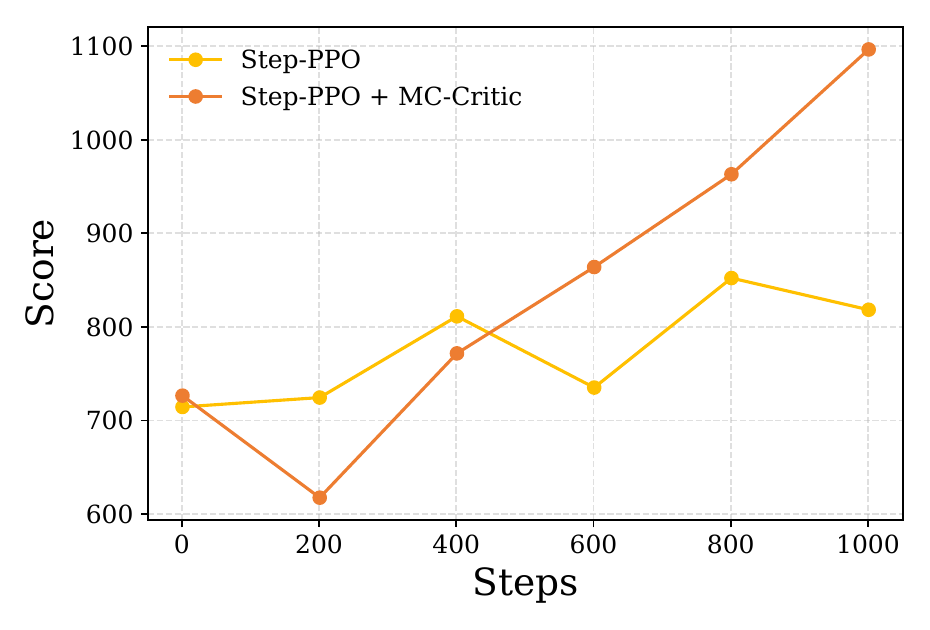}
        \caption{PPO on 2048}
        \label{fig:zero_ppo_2048}
    \end{subfigure}
    \hspace{2em} 
    \begin{subfigure}[b]{0.33\textwidth}
        \centering
        \includegraphics[width=\linewidth]{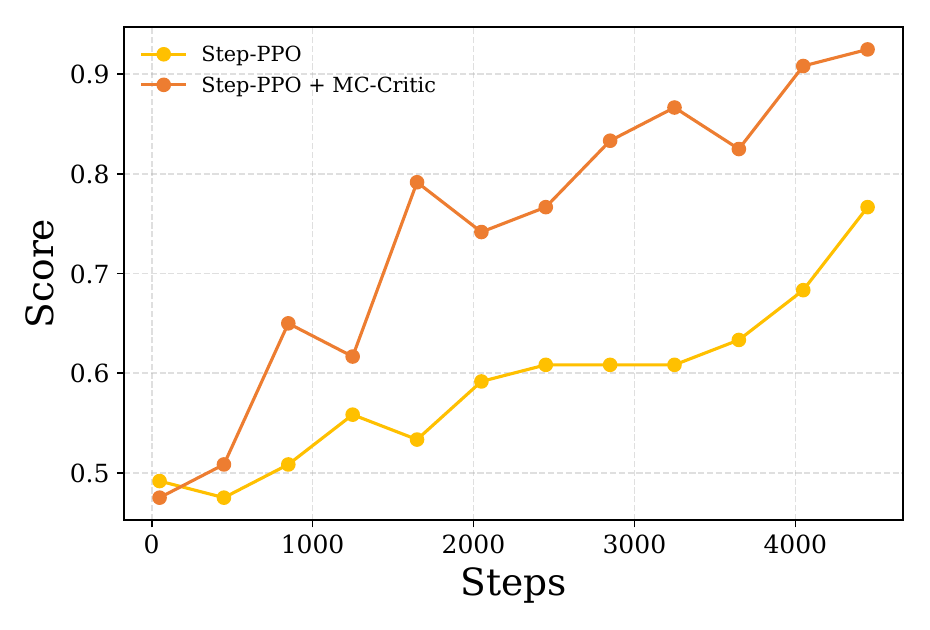}
        \caption{PPO on Sokoban}
        \label{fig:zero_ppo_sokoban}
    \end{subfigure}
    
    \par\vspace{1em} 
    
    \begin{subfigure}[b]{0.33\textwidth}
        \centering
        \includegraphics[width=\linewidth]{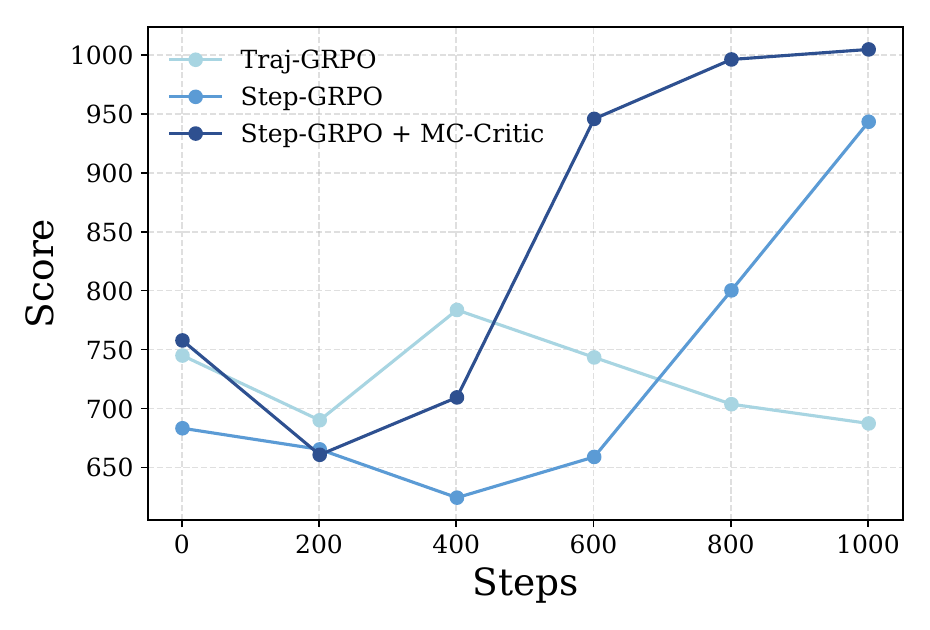}
        \caption{GRPO on 2048}
        \label{fig:zero_grpo_2048}
    \end{subfigure}
    \hspace{2em} 
    \begin{subfigure}[b]{0.33\textwidth}
        \centering
        \includegraphics[width=\linewidth]{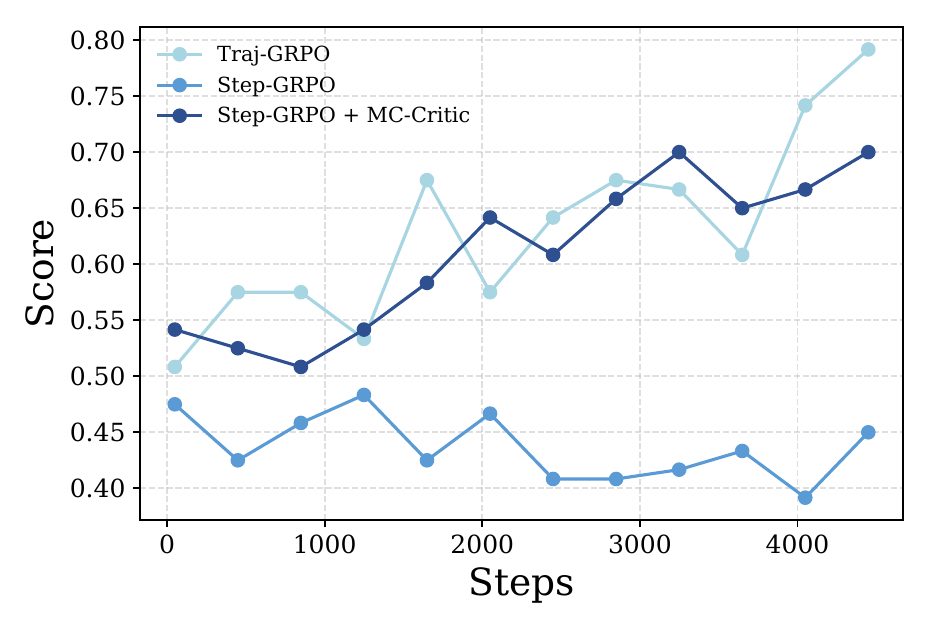}
        \caption{GRPO on Sokoban}
        \label{fig:zero_grpo_sokoban}
    \end{subfigure}
    
    \caption{Comparison of baseline RL methods and MC-Critic when trained from scratch on 2048 and Sokoban. MC-PPO and MC-GRPO are represented by ``Step-PPO + MC-Critic" and ``Step-GRPO + MC-Critic", respectively.}
    \label{fig:rl_zero}
    
\end{figure}

Figure~\ref{fig:rl_zero} reports results on 2048 under the standard setting and Sokoban on the training levels. Across both environments, MC-PPO achieves the highest scores, confirming the effectiveness of integrating MC-Critic into policy optimization. In 2048, MC-GRPO consistently outperforms other GRPO-based variants, further supporting the benefits of our approach. In Sokoban, MC-GRPO performs comparably to Traj-GRPO. This can be attributed to the simplicity of the training Sokoban levels: most levels are solved within a few steps, making entire trajectories sufficient as single training samples, without incurring significant variance.
However, as trajectory length increases, reward variance accumulates, destabilizing training for Traj-GRPO. This effect is evident in 2048, where Traj-GRPO performance degrades, whereas MC-GRPO remains stable and continues to improve, highlighting the robustness of MC-Critic in long-horizon tasks.

To evaluate generalization, we further test all methods on various environment variants. Table~\ref{tab:main_results} shows that MC-Critic outperforms the baseline methods on both 2048 and Sokoban variants, demonstrating its superior robustness and generalization.








\begin{table}[t]
\centering
\footnotesize
\setlength{\tabcolsep}{4pt}
\begin{tabular}{lccccc}
\toprule
 & \multicolumn{2}{c}{2048} & \multicolumn{3}{c}{Sokoban} \\
\cmidrule(lr){2-3} \cmidrule(lr){4-6}
Method & 3$\times$3 & 3072 & Extra & Action & Symbol \\
\midrule
Traj-GRPO & 202.0 & 1760.7 & 0.50 & 0.73 & 0.90 \\
Step-GRPO & 188.0 & 1792.9 & 0.25 & 0.53 & 0.86 \\
Step-PPO  & 202.7 & 1912.90 & 0.45 & 0.55 & 0.88 \\
\midrule
MC-GRPO   & 194.2 & 1754.7 & \textbf{0.55} & 0.80 & \textbf{1.03} \\
MC-PPO    & \textbf{239.8} & \textbf{2229.1} & 0.53 & \textbf{0.96} & 0.98 \\
\bottomrule
\end{tabular}
\caption{Performance of RL methods with and without MC-Critic on 2048 and Sokoban environment variants when trained from scratch.}
\label{tab:main_results}
\end{table}

\paragraph{Analysis.} 

\begin{figure}[h]
\centering

\begin{minipage}{0.48\textwidth}
    \centering
    \includegraphics[width=0.9\linewidth]{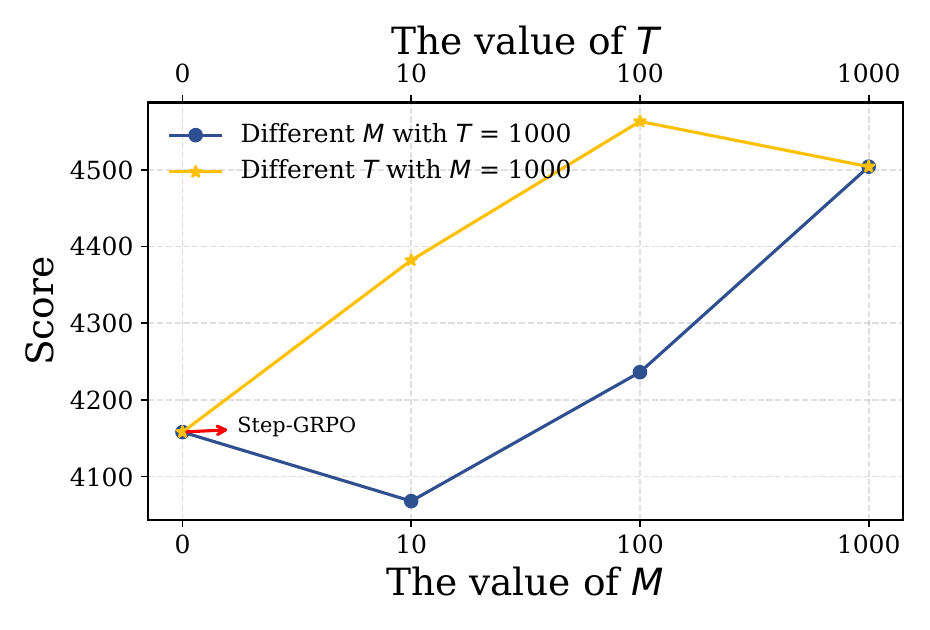}
    \caption*{(a) 2048}
\end{minipage}
\hfill
\begin{minipage}{0.48\textwidth}
    \centering
    \includegraphics[width=0.9\linewidth]{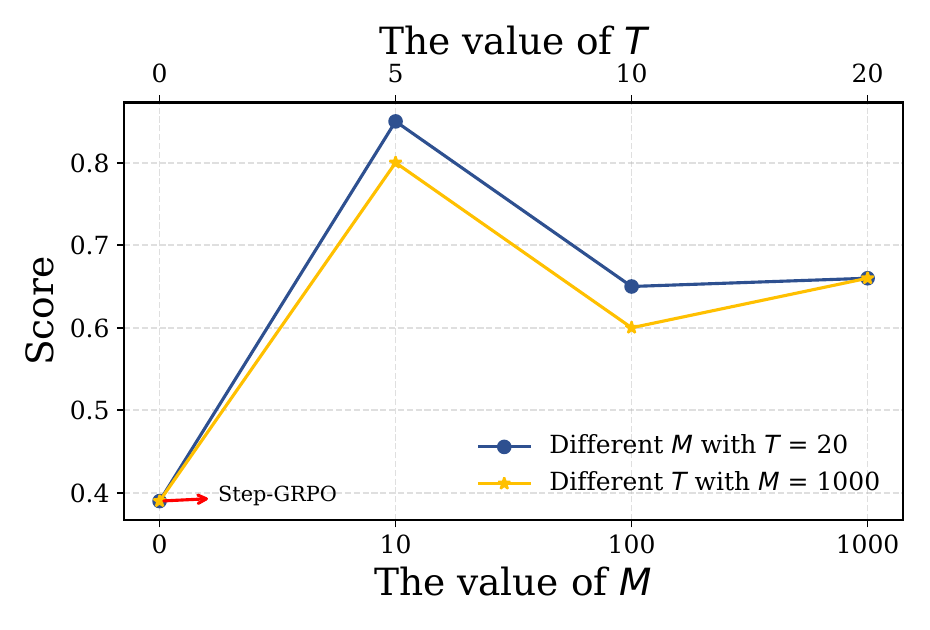}
    \caption*{(b) Sokoban}
\end{minipage}

\caption{The Performance of MC-GRPO with different number of trajectories ($M$) and maximum time steps within a trajectory ($T$) used in MC-Critic on the 2048 and Sokoban. When $M=0$ or $T=0$, MC-GRPO degenerates into Step-GRPO.}
\label{fig:exp_analysis}
\end{figure}

Finally, we analyze the impact of two critical hyperparameters within MC-Critic, as previously defined in Eq.~\ref{eq:mc_critic}: the number of trajectories, denoted by $M$, and the maximum number of steps per trajectory, denoted by $T$. We conduct analysis experiments employing MC-GRPO (i.e., Step-GRPO augmented with MC-Critic). 
To isolate the effects of varying $M$ and $T$, we perform an ablation study: we fix $M=1000$ while varying $T \in \{0, 10, 100, 1000\}$ for 2048 and $T \in \{0, 5, 10, 20\}$ for Sokoban, and conversely, fix $T=1000$ for 2048 and $T=20$ for Sokoban while varying $M \in \{0, 10, 100, 1000\}$. We train MC-GRPO under each configuration, with the final results summarized in Figure~\ref{fig:exp_analysis}. It is worth noting that when $M=0$ or $T=0$, the value term $V_{\pi_{random}}^{\text{MC}}(s_{t_u+1})$ in Eq.~\ref{eq:q_mc} becomes zero, resulting in $Q_{\pi_{random}}^{\text{MC}}(s_{t_u}, a_{t_u}^i) = r_{t_u}^i$. Consequently, MC-GRPO degenerates into Step-GRPO, the performance observed at $M=0$ or $T=0$ therefore represents the baseline performance of Step-GRPO.



We begin by analyzing the impact of $M$. On the 2048 environment, MC-GRPO with $M=10$ performs worse than Step-GRPO. This is attributed to the long horizon of the 2048 environment (which can involve hundreds of turns per trajectory). Averaging Monte-Carlo returns from only 10 trajectories yields a value estimate $V_{\pi_{random}}^{\text{MC}}$ with high variance—significantly higher than the variance associated with the immediate step-level rewards used in Step-GRPO. As we increase $M$ to 100 and 1000, the variance of $V_{\pi_{random}}^{\text{MC}}$ decreases monotonically, leading to improved performance. Furthermore, compared to using immediate step-level rewards in Step-GRPO, $V_{\pi_{random}}^{\text{MC}}$ accounts for the subsequent impact of current decisions across multiple turns, it facilitates the learning of multi-turn lookahead reasoning. This results in a substantial performance gain for MC-GRPO over Step-GRPO.

Conversely, on the Sokoban environment, a larger $M$ is not always better. Performance at $M=10$ surpasses that at $M=100$ and $1000$. We attribute this to the sparsity of rewards on Sokoban. Random policy rollouts rarely succeed, yielding zero Monte-Carlo returns for the majority of trajectories. Consequently, the positive Monte-Carlo returns from successful trajectories are diluted when averaged over $M$ trajectories. When $M$ is too large, the $Q_{\pi_{random}}^{\text{MC}}$ values for different actions become very close, obscuring the advantage of the optimal action. A smaller $M$, while having higher variance, preserves the distinction between $Q_{\pi_{random}}^{\text{MC}}$ values, helping the model distinguish and learn the optimal policy.

Next, we analyze the impact of $T$. On the 2048 environment, as $T$ increases, the agent gains a longer horizon, and performance improves effectively, saturating around $T=100$. Increasing $T$ further leads to a slight decline in performance due to the increased variance of single-trajectory Monte-Carlo returns. A similar trend is observed on Sokoban, where performance at $T=5$ exceeds that at $T=10$ and $20$. Since 5 steps are sufficient to solve most levels, further increasing $T$ tends to increase the variance of Monte-Carlo returns, thereby degrading performance.

The MC-Critic recipe can be summarized as follows: For environments with dense rewards, $M$ should be set as large as possible, provided that the interaction efficiency allows. For environments with sparse rewards, $M$ should not be excessive, as a smaller $M$ may yield better results. Additionally, $T$ should be set to the average number of steps required for a successful trajectory and should not be overly large.

\section{Conclusion}

In this work, we present ProAct, a comprehensive framework designed to empower LLM agents with accurate lookahead capabilities and stable policy optimization in long-horizon interactive environments. To address the critical challenges of compounding simulation errors and high-variance value estimation, ProAct introduces a two-stage training paradigm. First, Grounded Lookahead Distillation (GLAD) effectively bridges the gap between expensive inference-time search and efficient policy intuition by distilling explicit MCTS trajectories into concise, environment-grounded reasoning chains. Second, the Monte-Carlo Critic (MC-Critic) provides a low-variance, plug-and-play value estimator that significantly enhanced the stability of multi-turn agentic RL algorithms. Empirical results on distinct benchmarks—stochastic (2048) and deterministic (Sokoban)—demonstrate that a 4B parameter model trained with ProAct not only outperforms existing open-source baselines but also achieves strong generalization capabilities comparable to state-of-the-art proprietary models.

\bibliography{iclr2025_conference}
\bibliographystyle{iclr2025_conference}
\clearpage
\appendix
\clearpage
\section{Appendix}

\subsection{Environment and Evaluation Details}
\label{apend:env}

For both 2048 and Sokoban, the LLM agent interacts with the environment exclusively through textual observations and textual action outputs.
At each step, the agent is required to produce responses in the following structured format:

\begin{verbatim}
thought: <analysis and reasoning process>
action: <selected action>
\end{verbatim}

During execution, the environment wrapper parses the \texttt{action} field from the model output and applies it to the environment.
If the output does not follow the required format, or if the parsed action is invalid under the current state, the step is treated as an invalid action and penalized accordingly.
An episode terminates when a terminal state is reached or when the maximum trajectory length $T$ is exceeded.

\paragraph{2048.} 
We adopt the standard 2048 game rules.
The environment state is serialized as a textual representation of the board, printed in a list-like format with rows shown in order.
Empty cells are denoted by \texttt{.}.
An example observation is shown below:
\begin{verbatim}
{'board': '[128 8 4 2][ 8 4 . .][ 4 . . .][ 2 . . 2]'}
\end{verbatim}
At each step, the agent selects one of four actions: \textit{up}, \textit{down}, \textit{left}, or \textit{right}.
Whenever two tiles merge, the agent receives a step-level reward equal to the value of the resulting tile
(e.g., merging $2 + 2 \rightarrow 4$ yields a reward of $4$).
During evaluation, performance is measured by the cumulative merge score obtained when the episode terminates.
The same evaluation protocol is applied to all environment variants, including the $3 \times 3$ grid and the 3072 setting.

\paragraph{Sokoban.}
In Sokoban, the environment presents a symbolic ASCII map to the agent, where different characters represent walls, boxes, target locations, the agent, and empty cells.
Table~\ref{tab:sokoban_symbol} summarizes the symbol mappings for the standard environment and the symbolic-variant setting.
Below is an example Sokoban observation from the SFT evaluation levels:
\begin{verbatim}
Current Sokoban board state:
#########
## *    #
### ?   #
######  #
######$ #
##### @##
#####   #
######  #
#### #  #
###     #
#########

Briefly analyze the board and provide your move decision.
\end{verbatim}
The agent can move using four directional actions: \textit{up}, \textit{down}, \textit{left}, and \textit{right}.
When adjacent to a box, the agent may execute a push action by prefixing the direction with \texttt{push}
(e.g., \textit{push up}).
For evaluation, we report the average number of boxes successfully pushed onto target locations per level.
This metric provides a smoother signal than binary success, as the number of boxes per level varies from 1 to 5.

For RL training from scratch, we find that Sokoban levels used in SFT training are overly challenging and lead to insufficient exploration.
We therefore construct a set of simplified Sokoban levels that can be solved within 20 steps.
Two example levels are shown below:
\begin{center}
\begin{minipage}{0.45\linewidth}
\begin{verbatim}
#####
#@  #
#$ $#
#? ?#
#####
\end{verbatim}
\end{minipage}
\hfill
\begin{minipage}{0.45\linewidth}
\begin{verbatim}
#####
#@  #
# $ #
#  ?#
#####
\end{verbatim}
\end{minipage}
\end{center}

Accordingly, the maximum trajectory length is set to $T=20$ for this setting.
These simplified levels are used only for RL training from scratch; evaluation is still conducted on the original test distributions unless otherwise specified.

\begin{table}[t]
\centering
\caption{Symbol mappings used in the standard Sokoban environment and the symbolic-variant environment.}
\label{tab:sokoban_symbol}
\begin{tabular}{l c c}
\toprule
Semantic Meaning & Standard Symbol & Symbol Variant \\
\midrule
Wall             & \texttt{\#} & \texttt{\#} \\
Floor            & \texttt{ }  & \texttt{\_} \\
Target           & \texttt{?}  & \texttt{O} \\
Box on target    & \texttt{*}  & $\sqrt{}$ \\
Box              & \texttt{\$} & \texttt{B} \\
Player           & \texttt{@}  & \texttt{P} \\
Player on target & \texttt{+}  & \texttt{+} \\
\bottomrule
\end{tabular}
\end{table}

For Sokoban evaluation, performance is measured by the average number of boxes successfully placed onto target locations per level, averaged over multiple independent runs.
Each run is capped at a maximum of 200 steps, after which the episode is terminated.
To ensure stable scoring when a run ends due to the step limit, we record the maximum number of boxes placed on targets at any point during the episode, rather than the final configuration.
In addition, to reduce evaluation time, episodes are terminated early when the environment is detected to be in a deadlock state, where no further progress is possible.

\subsection{Training Config}
\label{apend:train_config}

In this section, we detail the training configurations used in SFT and RL experiments, which are presented in Table~\ref{tab:sft_config}-~\ref{tab:rl_zero_config}.

\begin{table}[htbp!]
    \centering
    \caption{Training configurations used in SFT experiments.}
    \label{tab:sft_config}
    \begin{tabular}{lcc}
        \toprule
        \textbf{Hyperparameter} & \textbf{2048} & \textbf{Sokoban} \\
        \midrule
        Learning Rate & $5 \times 10^{-5}$ & $2 \times 10^{-5}$ \\
        Batch Size & 16 & 16 \\
        Training Epochs & 4 & 10 \\
        Optimizer & Adam & Adam \\
        \bottomrule
    \end{tabular}
\end{table}

\begin{table}[htbp!]
    \centering
    \caption{Training configurations used in RL experiments with GLAD supervision.}
    \label{tab:rl_sft_config}
    \begin{tabular}{lcc}
        \toprule
        \textbf{Hyperparameter} & \textbf{2048} & \textbf{Sokoban} \\
        \midrule
        Actor Learning Rate for PPO/GRPO & $1 \times 10^{-6}$ & $1 \times 10^{-6}$ \\
        Critic Learning Rate for PPO & $1 \times 10^{-5}$ & $1 \times 10^{-5}$ \\
        Group Size in GRPO & 8 & 8 \\
        Batch Size in PPO/GRPO & 32 & 32 \\
        Optimizer & Adam & Adam \\
        Training Epochs Per Sample in GRPO & 10 & 1 \\
        Training Epochs Per Sample in PPO & 1 & 1 \\
        $M$ for MC-Critic & 1000 & 1000 \\
        $T$ for MC-Critic & 1000 & 200 \\
        $\gamma$ for Calculating Monte-Carlo Return & 0.9 & 1 \\
        $\epsilon$ for Ratio Clipping in PPO/GRPO & 0.4 & 0.4 \\
        \bottomrule
    \end{tabular}
\end{table}

\begin{table}[htbp!]
    \centering
    \caption{Training configurations used in RL experiments without GLAD supervision.}
    \label{tab:rl_zero_config}
    \begin{tabular}{lcc}
        \toprule
        \textbf{Hyperparameter} & \textbf{2048} & \textbf{Sokoban} \\
        \midrule
        Actor Learning Rate for PPO/GRPO & $1 \times 10^{-6}$ & $1 \times 10^{-6}$ \\
        Critic Learning Rate for PPO & $1 \times 10^{-5}$ & $1 \times 10^{-5}$ \\
        Group Size in GRPO & 8 & 4 \\
        Batch Size in PPO/GRPO & 16 & 16 \\
        Optimizer & Adam & Adam \\
        Training Epochs Per Sample in GRPO & 1 & 1 \\
        Training Epochs Per Sample in PPO & 1 & 1 \\
        $M$ for MC-Critic & 1000 & 1000 \\
        $T$ for MC-Critic & 1000 & 20 \\
        $\gamma$ for Calculating Monte-Carlo Return & 0.9 & 1 \\
        $\epsilon$ for Ratio Clipping in PPO/GRPO & 0.4 & 0.4 \\
        \bottomrule
    \end{tabular}
\end{table}

\newpage

\subsection{Prompt Template and Example Outputs}
\label{apend:prompt}

This section presents the full prompt used in the case study, together with two representative model outputs. The first example is generated by the baseline model (Qwen3-4B-Instruct), while the second is produced by the same model after supervised training with GLAD.

\begin{tcolorbox}[
    colback=gray!10,       
    colframe=black,        
    colbacktitle=black,    
    title=\textbf{Qwen3-4B-Instruct}, 
    fonttitle=\bfseries\Large,
    boxrule=1pt,
    width=\textwidth,
    enhanced,
    sharp corners,
    breakable,
    listing only,
    listing options={basicstyle=\ttfamily\small}
]
    \textbf{\textcolor{blue}{System Prompt:}} \\
    You are an intelligent 2048 game player. Briefly analyze the board state and provide your reasoning process and move decision.

Response format:

thought: [your analysis and reasoning process]

move: [up/down/left/right]

Strategy tips:
- Keep the highest tile in a corner
- Prioritize creating merge opportunities and empty spaces
- Consider chain reactions after each move

    \vspace{1em} 
    
    \textbf{\textcolor{blue}{User Prompt:}} \\
    Current 2048 board state:
\{`board': `[  128     8     4     2]\textbackslash n[    8     4     .     .]\textbackslash n[    4     .     .     .]\textbackslash n[    2     .     .     2]'\}

Briefly analyze the board and provide your move decision.

\vspace{1em} 

\textbf{\textcolor{blue}{Response:}} \\
thought: The board has a 128 in the top-left corner, which is a strong position to keep. The 8 and 4 tiles are already aligned in a way that could potentially merge if we create a path. The empty spaces (denoted by '.') are mostly on the right and bottom, suggesting we should focus on expanding from the left and top to create merge opportunities. A left move would push the 128, 8, 4, 2 to the left, but that might not help. A down move could allow the 2 at the bottom-right to merge with the 2 on the bottom-left, but that's only 2+2=4. However, a better move is to move down — this will create a space in the top row and allow the 8 and 4 to potentially align with the 128 in the future. But actually, the best move is to move **left**, which will shift the top row left and allow the 8 and 4 to align with the 128, creating a path for merging. However, the 8 and 4 are already to the right of 128 — so moving left will push them leftward, possibly creating a merge with 128. But 128 is already at the corner. Actually, moving **left** will cause the 128 to stay, and the 8 and 4 to shift left, possibly creating a merge between 8 and 4 into 16, then 16 and 128 into 256. But currently, the 8 and 4 are not adjacent. After checking positions:

- Row 0: [128, 8, 4, 2]
- Row 1: [8, 4, ., .]
- Row 2: [4, ., ., .]
- Row 3: [2, ., ., 2]

A left move will shift all tiles left, so row 0 becomes [128, 8, 4, 2] → [128, 8, 4, 2] (no change since already left-aligned), row 1 → [8, 4, ., .], row 2 → [4, ., ., .], row 3 → [2, ., ., 2] → [2, ., ., 2] (no change). So no merge.

A down move: pushes each row down. Row 0: [128,8,4,2] → [.,.,.,.] → then new row 0 becomes [.,.,.,.] and the rest shift down. But actually, down move will push tiles down, so:

- Row 0: [128,8,4,2] → moves down → row 1: [8,4,.,.] → row 2: [4,.,.,.] → row 3: [2,.,.,2] → then new row 0 becomes [128,8,4,2] → no, actually in 2048, tiles fall down and merge.

Actually, correct logic: when moving down, each column collapses down. So:

Column 0: 128, 8, 4, 2 → becomes [128,8,4,2] → after falling, it's [128,8,4,2] → no merge because not adjacent.

Column 1: 8,4,.,. → becomes [8,4,.,.] → then 8 and 4 can merge? No, not adjacent.

Wait — in 2048, when moving down, tiles fall down and merge if adjacent in the same column.

So column 0: 128,8,4,2 → all different → no merge.

Column 1: 8,4,.,. → becomes [8,4,.,.] → no merge.

Column 2: 4,.,.,. → becomes [4,.,.,.] → no merge.

Column 3: 2,.,.,2 → becomes [2,.,.,2] → no merge.

No merges.

Now try **right** move: tiles shift right.

Row 0: [128,8,4,2] → stays same (already right-aligned)

Row 1: [8,4,.,.] → becomes [.,.,8,4]

Row 2: [4,.,.,.] → becomes [.,.,.,4]

Row 3: [2,.,.,2] → becomes [.,.,.,2]

No merges.

Now **up** move: tiles fall up.

Column 0: 128,8,4,2 → becomes [128,8,4,2] → no merge.

Column 1: 8,4,.,. → becomes [8,4,.,.] → no merge.

Column 2: 4,.,.,. → becomes [4,.,.,.] → no merge.

Column 3: 2,.,.,2 → becomes [2,.,.,2] → no merge.

Still no merges.

But the key is to create a merge path. The 128 is in the corner — good. The 8 and 4 in row 0 can merge into 16 if they are adjacent. Currently, 8 and 4 are adjacent — 8 and 4 → can merge? Only if they are equal. 8 and 4 are not equal. So no.

But if we move **left**, the 8 and 4 in row 0 will shift left — but they are already left of 128. So no.

Wait — perhaps the best move is **down** to create space in the top row for future merges.

But actually, the most promising is to move **left** to push the 8 and 4 toward the 128 — but they are already next to it.

Wait — row 0: [128, 8, 4, 2] — so 128 is at index 0, then 8 at 1, 4 at 2, 2 at 3.

If we move left, all tiles shift left — so row 0 becomes [128,8,4,2] → no change.

But row 1: [8,4,.,.] → becomes [8,4,.,.] → no change.

Row 2: [4,.,.,.] → becomes [4,.,.,.]

Row 3: [2,.,.,2] → becomes [2,.,.,2]

No merges.

But what if we move **down**? Then the tiles fall down, and we might create a new alignment.

After down move:

- Column 0: 128,8,4,2 → becomes [128,8,4,2] → no merge
- Column 1: 8,4,.,. → becomes [8,4,.,.] → no merge
- Column 2: 4,.,.,. → becomes [4,.,.,.] → no merge
- Column 3: 2,.,.,2 → becomes [2,.,.,2] → no merge

Still no.

Wait — perhaps the best move is to move **right** to create space on the right for future tiles.

But the real opportunity: the 8 in row 0, col 1 and 8 in row 1, col 0 — not aligned.

Alternatively, consider that 4 in row 0 col 2 and 4 in row 1 col 1 — not adjacent.

But after a **left** move, the 8 and 4 in row 0 are still not adjacent to 128.

Wait — 128 is at (0,0), 8 at (0,1), so they are adjacent — but 128 and 8 are not equal — cannot merge.

So no merge.

The only possible merge is if two 4s or two 8s come together.

Currently, we have:
- 128, 8, 4, 2 in row 0
- 8, 4 in row 1
- 4 in row 2
- 2, 2 in row 3

So 8 appears at (0,1), (1,0) — not adjacent
4 appears at (0,2), (1,1), (2,0) — not adjacent
2 appears at (0,3), (3,0), (3,3)

So no adjacent duplicates.

But if we move **down**, we create space at the top, and the 2 at (3,3) might fall into (2,3), etc.

But still, no merge.

Best strategy: create a path to merge two 4s or two 8s.

Currently, the 4 at (0,2) and (1,1) are diagonal — not adjacent.

The 4 at (2,0) is isolated.

The 8 at (0,1) and (1,0) are diagonal.

So no merge possible.

But the board is not dead — we can still make progress.

Best move: **down** — to create space in the top row for future tiles to fall into.

Alternatively, **left** — but no change.

Wait — in 2048, when you move, tiles fall and merge if adjacent and equal.

So if we move **down**, the tiles fall down, so:

- Column 0: 128,8,4,2 → after falling down, it becomes [128,8,4,2] — no merge
- Column 1: 8,4,.,. → becomes [8,4,.,.] — no merge
- Column 2: 4,.,.,. → becomes [4,.,.,.] — no merge
- Column 3: 2,.,.,2 → becomes [2,.,.,2] — no merge

No merge.

But if we move **right**, tiles shift right:

- Row 0: [128,8,4,2] → becomes [.,.,.,.] — no, wait: shifting right means each row shifts right, so empty spaces are filled on left.

So row 0: [128,8,4,2] → becomes [.,.,.,.] — no, actually in 2048, when you move right, tiles fall to the right.

So row 0: [128,8,4,2] → already at right — stays.

Row 1: [8,4,.,.] → becomes [.,.,8,4]

Row 2: [4,.,.,.] → becomes [.,.,.,4]

Row 3: [2,.,.,2] → becomes [.,.,.,2]

Now, in column 3: we have 2 (from row0), 4 (row1), 4 (row2), 2 (row3) — no, row0 col3 is 2, row1 col3 is 4, row2 col3 is 4, row3 col3 is 2 — so column 3: [2,4,4,2] → after falling right, it becomes [.,.,.,2] → no, after right move, each row shifts right, so:

After right move:
- Row 0: [128,8,4,2] → stays
- Row 1: [.,.,8,4]
- Row 2: [.,.,.,4]
- Row 3: [.,.,.,2]

So column 3: [2,4,4,2] — but after shift, they are at the right — so in column 3: row0:2, row1:4, row2:4, row3:2 — now, in the same column, row1 and row2 both have 4 — so they are adjacent vertically.

After the right move, the board becomes:

[128,8,4,2]
[.,.,8,4]
[.,.,.,4]
[.,.,.,2]

Now, in column 3: row1 has 4, row2 has 4 — they are adjacent vertically — so they will merge into 8 when the move is applied.

But only if the move is **down** — no, after a right move, the tiles are shifted, and then if there's a merge, it happens in the same move.

In the right move, tiles fall to the right, and merge if adjacent and equal.

So in column 3: row1 has 4, row2 has 4 — they are in the same column, adjacent — so they merge into 8.

So after right move:

- Column 3: row1 and row2 both have 4 → merge into 8 at row1 or row2?

In 2048, when merging, the larger tile stays and the smaller is removed.

So in column 3: [2,4,4,2] → after right move, the 4 and 4 merge into 8, and the 2s remain.

But in the same column, after shifting right, the values are:

- Row0: col3 = 2
- Row1: col3 = 4
- Row2: col3 = 4
- Row3: col3 = 2

After right move, tiles fall to the right — so they are already at the right.

Then, in column 3, row1 and row2 have 4 — they are adjacent — so they merge into 8.

So the new board after right move:

- Row0: [128,8,4,2]
- Row1: [.,.,.,8]  (after merge, the 4 and 4 become 8 at row1 or row2?)
- Actually, in column 3: row1 and row2 both have 4 → they merge into one 8 at row1 or row2? Usually, the lower one wins, or the one that is in the middle.

Standard rule: tiles merge when adjacent and equal, and the value doubles.

So in column 3: row1 and row2 have 4 — they merge into 8 at row1 (or row2) — say row1 gets 8, row2 becomes empty.

So after merge:

- Row0: [128,8,4,2]
- Row1: [.,.,.,8]
- Row2: [.,.,.,.] (since row2 col3 was 4, now gone)
- Row3: [.,.,.,2]

And a new tile appears somewhere — usually in a random empty cell.

But the key point is that a merge occurs — creating a 8 from two 4s.

So a **right** move creates a merge.

Similarly, a down move might not.

But in this case, right move creates a merge.

Also, the 128 is in the corner — good.

So the best move is **right** to create a merge of two 4s into 8.

Therefore, move: right

But wait — in the original board, row1 col3 and row2 col3 are both empty — after right move, row1 col3 and row2 col3 are filled with 4 and 4 — yes, from the original row1: [8,4,.,.] → after right move, becomes [.,.,8,4] — so row1 col3 = 4

Row2: [4,.,.,.] → becomes [.,.,.,4] — row2 col3 = 4

So yes, in column 3, row1 and row2 have 4 — adjacent — merge into 8.

So merge occurs.

Thus, this is a valid and productive move.

Therefore, the best move is **right**.

move: right
\end{tcolorbox}

\begin{tcolorbox}[
    colback=gray!10,       
    colframe=black,        
    colbacktitle=black,    
    title=\textbf{Qwen3-4B-Instruct + GLAD}, 
    fonttitle=\bfseries\Large,
    boxrule=1pt,
    width=\textwidth,
    enhanced,
    sharp corners,
    breakable,
    listing only,
    listing options={basicstyle=\ttfamily\small}
]
    \textbf{\textcolor{blue}{System Prompt:}} \\
    You are an intelligent 2048 game player. Briefly analyze the board state and provide your reasoning process and move decision.

Response format:

thought: [your analysis and reasoning process]

move: [up/down/left/right]

Strategy tips:
- Keep the highest tile in a corner
- Prioritize creating merge opportunities and empty spaces
- Consider chain reactions after each move

    \vspace{1em} 
    
    \textbf{\textcolor{blue}{User Prompt:}} \\
    Current 2048 board state:
\{`board': `[  128     8     4     2]\textbackslash n[    8     4     .     .]\textbackslash n[    4     .     .     .]\textbackslash n[    2     .     .     2]'\}

Briefly analyze the board and provide your move decision.

\vspace{1em} 

\textbf{\textcolor{blue}{Response:}} \\
thought: Let me think. The 128 is anchored top-left, and the top row is nicely monotonic as 128-8-4-2. There are 7 empty cells, so the position is flexible. If I go up, the right column [2, ., ., 2] compresses and the two 2s merge into a 4 at the top, turning the top row into 128-8-4-4. That increases empties from 7 to 8 and keeps the 128 in the corner, which is good. It also sets up a likely follow-up left to combine the two 4s into an 8, strengthening the row.

If I go left, the bottom row [2, ., ., 2] becomes [2, 2, ., .] and merges into a 4, making column 1 [128, 8, 4, 4]. Empties also go to 8. This is fine and sets up a future up merge in column 1, but it doesn’t improve the top row immediately.

If I go down, the right column’s 2s merge at the bottom, but column 2 [8, 4, ., .] drops to [., ., 8, 4], pulling small tiles away from the top row and weakening alignment.

Right similarly merges bottom into a 4 on the far right, but pushes lower tiles rightward, which is less consistent with a top-left anchor.

Given these trade-offs, up is a better choice because it creates the 4-4 on the top row, adds one empty space, and preserves the corner strategy.

move: up

\end{tcolorbox}


\end{document}